\documentclass[runningheads]{llncs}
\usepackage{graphicx}
\usepackage{comment}
\usepackage{amsmath,amssymb} 
\usepackage{color}

\usepackage{graphicx}
\usepackage{amsmath}
\mathchardef\mhyphen="2D 
\usepackage{amssymb}
\usepackage{enumitem}
\usepackage[numbers]{natbib}
\usepackage{amsmath}
\usepackage{siunitx}
\usepackage{booktabs}
\usepackage{subcaption}
\usepackage{caption}
\usepackage{hhline}
\usepackage{comment}
\usepackage{floatrow}
\usepackage{hyperref}
\newfloatcommand{capbtabbox}{table}[][\FBwidth]

\newcommand{\METHOD}{ScanRefer}
\newcommand{\DATASET}{ScanRefer}
\newcommand{\NUMSCENES}{800}

\newcommand{\NUMOBJECTS}{11,046}
\newcommand{\NUMOBJECTSPS}{13.81}
\newcommand{\NUMDESCSPS}{64.48}
\newcommand{\AVGDESCS}{4.67}
\newcommand{\NUMDESCS}{51,583}
\newcommand{\NUMTRAIN}{36,665}
\newcommand{\NUMVAL}{9,508}
\newcommand{\NUMTEST}{5,410}
\newcommand{\NUMUNIQUE}{1,875}
\newcommand{\NUMMULTIPLE}{7,663}

\newcommand{\AVGLEN}{20.27}
\newcommand{\VOCABSIZE}{4,197}

\newcommand{\mypara}[1]{\noindent\textbf{#1}}

\definecolor{orange}{rgb}{1.0, 0.49, 0.0}

\newcommand{\updated}[1]{{#1}}
\newcommand{\red}[1]{{\color{red}{#1}}}
\newcommand{\green}[1]{{\color{green}{#1}}}
\newcommand{\blue}[1]{{\color{blue}{#1}}}
\newcommand{\orange}[1]{{\color{orange}{#1}}}

\begin{document}
\pagestyle{headings}
\mainmatter
\def\ECCVSubNumber{3408}  

\title{\METHOD: 3D Object Localization in RGB-D Scans using Natural Language} 

\titlerunning{\METHOD: 3D Object Localization in RGB-D Scans using Natural Language}
\authorrunning{Chen et al.} 
\author{
Dave Zhenyu Chen$^{1}$ \qquad \qquad  Angel X. Chang$^{2}$ \qquad \qquad Matthias Nie{\ss}ner$^{1}$ \\
\qquad \\
$^{1}$Technical University of Munich \qquad $^{2}$Simon Fraser University \\
}
\institute{}

\maketitle

\begin{figure}
    \centering
    \includegraphics[width=0.9\linewidth]{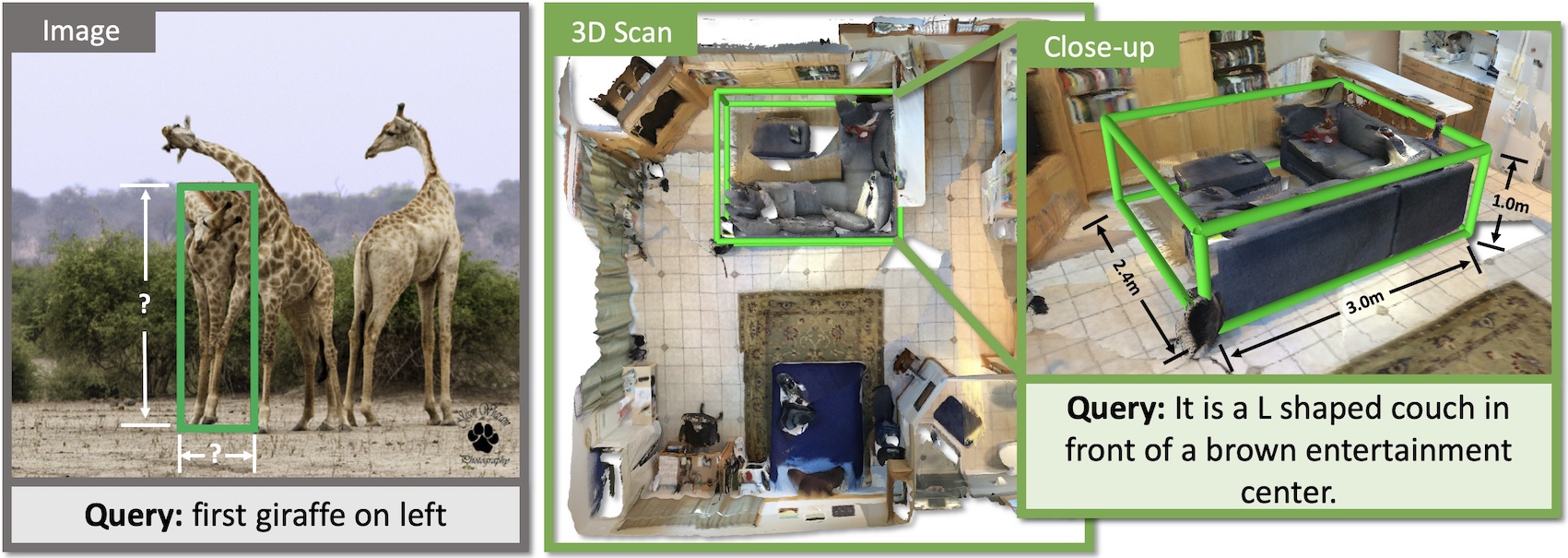}
    \caption{
    We introduce the task of object localization in 3D scenes using natural language. Given as input a 3D scene and a natural language expression, we predict the bounding box for the target 3D object (right).
    The counterpart 2D task (left) does not capture the physical extent of the 3D objects.
    }
    \label{fig:2d_vs_3d}
\end{figure}



\begin{abstract}

We introduce the task of 3D object localization in RGB-D scans using natural language descriptions.
As input, we assume a point cloud of a scanned 3D scene along with a free-form description of a specified target object.
To address this task, we propose \textbf{\METHOD}, learning a fused descriptor from 3D object proposals and encoded sentence embeddings.
This fused descriptor correlates language expressions with geometric features, enabling regression of the 3D bounding box of a target object.
We also introduce the \text{\DATASET} dataset, containing $\NUMDESCS$ descriptions of $\NUMOBJECTS$ objects from $\NUMSCENES$ ScanNet \citep{dai2017scannet} scenes. 
\text{\DATASET} is the first large-scale effort to perform object localization via natural language expression directly in 3D \footnote{Project page: \url{https://daveredrum.github.io/ScanRefer/}}.

\end{abstract}


\section{Introduction}

In recent years, there has been tremendous progress in both semantic understanding and localization of objects in 2D images from natural language (also known as visual grounding).  
Datasets such as ReferIt~\cite{kazemzadeh2014referitgame}, RefCOCO~\cite{yu2016modeling}, and Flickr30K Entities~\cite{plummer2015flickr30k} have enabled the development of various methods for visual grounding in 2D~\cite{hu2016natural,hu2016segmentation,mao2016generation}.
However, these methods and datasets are restricted to 2D images, where object localization fails to capture the true 3D extent of an object (see Fig.~\ref{fig:2d_vs_3d}, left).
This is a limitation for applications ranging from assistive robots to AR/VR agents where understanding the global 3D context and the physical size is important, e.g., finding objects in large spaces, interacting with them, and understanding their spatial relationships.
Early work by Kong et al.~\citep{kong2014you} looked at coreference in 3D, but was limited to single-view RGB-D images.

In this work, we address these shortcomings by proposing the task of object localization using natural language directly in 3D space.
Specifically, we develop a neural network architecture that localizes objects in 3D point clouds given natural language descriptions referring to the underlying objects; i.e., for a given text description in a 3D scene, we predict a corresponding 3D bounding box matching the best-described object.
To facilitate the task, we collect the \text{\DATASET} dataset, which provides natural language descriptions for RGB-D scans in ScanNet~\cite{dai2017scannet}.
In total, we acquire $\NUMDESCS$ descriptions of $\NUMOBJECTS$ objects. 
To the best of our knowledge, our \text{\DATASET} dataset is the first large-scale effort that combines 3D scene semantics and free-form descriptions.
In summary, our contributions are as follows:
\begin{itemize}[noitemsep]
\item We introduce the task of localizing objects in 3D environments using natural language descriptions.
\item We provide the \text{\DATASET} dataset containing $\NUMDESCS$ human-written free-form descriptions of $\NUMOBJECTS$ objects in 3D scans.
\item We propose a neural network architecture for localization based on language descriptions that directly fuses features from 2D images and language expressions with 3D point cloud features.
\item We show that our end-to-end method outperforms the best 2D visual grounding method that simply backprojects its 2D predictions to 3D by a significant margin (9.04 Acc@0.5IoU vs. \updated{22.39} Acc@0.5IoU).

\end{itemize}


\section{Related Work}

\mypara{Grounding Referring Expressions in Images.}
There has been much work connecting images to natural language descriptions across tasks such as image captioning~\citep{karpathy2014deep, karpathy2015deep, vinyals2015show, xu2015show}, text-to-image retrieval~\citep{wang2016learning,huang2018learning}, and visual grounding~\citep{hu2016natural,mao2016generation,yu2018mattnet}.
The task of visual grounding (with variants also known as referring expression comprehension or phrase localization) is to localize a region described by a given referring expression, the query.
Localization can be specified by a 2D bounding box~\cite{kazemzadeh2014referitgame,plummer2015flickr30k,mao2016generation} or a segmentation mask~\cite{hu2016segmentation}, with the input description being short phrases~\cite{kazemzadeh2014referitgame,plummer2015flickr30k} or more complex descriptions~\cite{mao2016generation}.
Recently, Acharya et al~\cite{acharya2019vqd} proposed visual query detection where the input is a question.
The focus of our work is to lift this task to 3D, focusing on complex descriptions that can localize an unique object in a scene.

Existing methods focus on predicting 2D bounding boxes~\cite{hu2016natural,rohrbach2016grounding,wang2016learning,wang2018learning,plummer2018conditional,yu2016modeling,yu2018mattnet,dogan2019neural,liu2019improving} and some predict segmentation masks~\cite{hu2016segmentation, liu2017recurrent, li2018referring, margffoy2018dynamic,ye2019cross,chen2019see}.
A two-stage pipeline is common, where first an object detector, either unsupervised~\cite{zitnick2014edge} or pretrained~\cite{ren2015faster}, is used to propose regions of interest, and then the regions are ranked by similarity to the query, with the highest scoring region provided as the final output.
Other methods address the referring expression task with a single stage end-to-end network~\cite{hu2016segmentation,nguyen2018object,yang2019fast}.
\updated{
There are also approaches that incorporate syntax~\cite{liu2019learning,hong2019learning}, use graph attention networks~\cite{wang2019neighbourhood,yang2019cross,yang2019dynamic}, speaker-listener models~\cite{mao2016generation,yu2017joint}, weakly supervised methods~\cite{xiao2017weakly,zhao2018weakly,datta2019align2ground} or tackle zero-shot settings for unseen nouns~\cite{sadhu2019zero}.}

However, all these methods operate on 2D image datasets~\citep{plummer2015flickr30k, kazemzadeh2014referitgame, yu2016modeling}.
A recent dataset~\citep{mauceri2019sun} integrates RGB-D images but lacks the complete 3D context beyond a single image.
Qi et al.~\citep{qi2020reverie} study referring expressions in an embodied setting, where semantic annotations are projected from 3D to 2D bounding boxes on images observed by an agent.
Our contribution is to lift NLP tasks to 3D by introducing the first large-scale effort that couples free-form descriptions to objects in 3D scans.
Tab.~\ref{tab:comp_datasets} summarizes the difference between our ScanRefer dataset and existing 2D datasets.

\begin{table}[t]
    \caption{Comparison of referring expression datasets in terms of the number of objects (\#objects), number of expressions (\#expressions), average lengths of the expressions, data format and the 3D context.}
    \label{tab:comp_datasets}
    \resizebox{\columnwidth}{!}{
        \begin{tabular}{l r r r c c}
            \toprule
            dataset & \#objects & \#expressions & AvgLeng & data format & 3D context \\
            \hline
            ReferIt~\citep{kazemzadeh2014referitgame} & 96,654 & 130,364 & 3.51 & image & - \\
            RefCOCO~\citep{yu2016modeling} & 50,000 & 142,209 & 3.50 & image & - \\
            Google RefExp~\citep{mao2016generation} & 49,820 & 95,010 & 8.40 & image & - \\
            SUN-Spot~\citep{mauceri2019sun} & 3,245 & 7,990 & 14.04 & image & depth \\
            REVERIE~\citep{qi2020reverie} & 4,140 & 21,702 & 18.00 & image & panoramic image \\
            \textbf{ScanRefer (ours)} & \textbf{\NUMOBJECTS} & \textbf{\NUMDESCS} & \textbf{\AVGLEN} & \textbf{3D scan} & \textbf{depth, size, location, etc.} \\
            \bottomrule
        \end{tabular}
    }
\end{table}

\mypara{Object Detection in 3D.}
Recent work on 3D object detection on volumetric grids~\citep{hou20193dsis, hou20193dsic, lahoud20193d, narita2019panopticfusion, elich20193d} has been applied to several 3D RGB-D datasets~\citep{song2015sun, dai2017scannet, chang2017matterport3d}.
As an alternative to regular grids, point-based methods, such as PointNet~\citep{qi2017pointnet} or PointNet++~\citep{qi2017pointnet++}, have been used as backbones for 3D detection and/or object instance segmentation~\citep{yang2019learning, engelmann2019dilated}. 
Recently, Qi et al.~\citep{qi2019deep} introduced VoteNet, a 3D object detection method for point clouds based on Hough Voting~\citep{hough1959machine}.
Our approach extracts geometric features in a similar fashion, but backprojects 2D feature information since the color signal is useful for describing 3D objects with natural language.

\begin{figure}[t!]
    \centering
    \includegraphics[width=\linewidth]{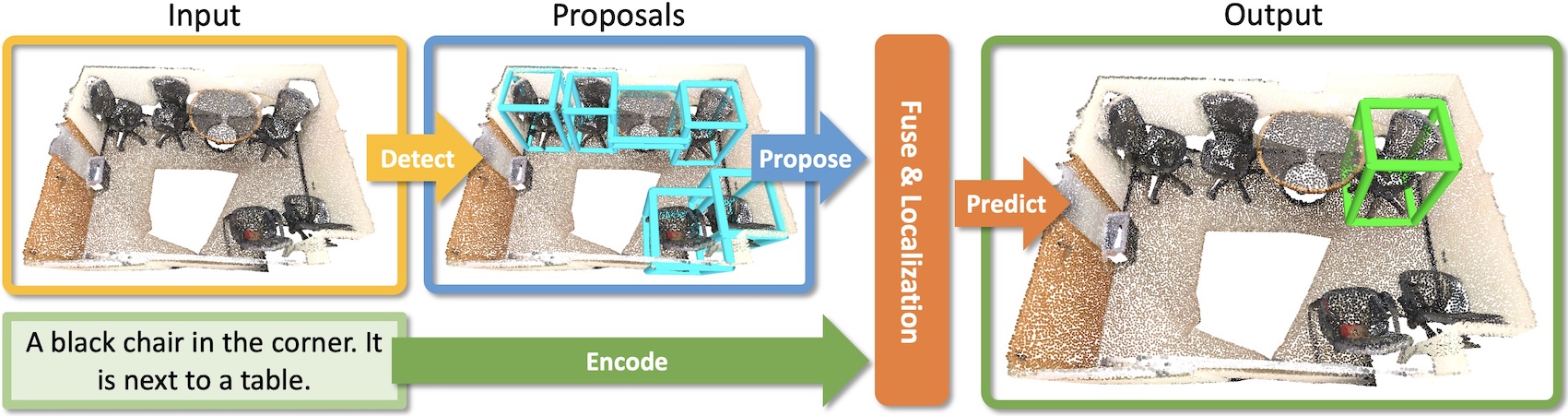}
    \caption{Our task: \text{\METHOD} takes as input a 3D scene point cloud and a description of an object in the scene, and predicts the object bounding box.}
    \label{fig:overview}
\end{figure}

\begin{figure}[t!]
    \centering
    \includegraphics[width=\linewidth]{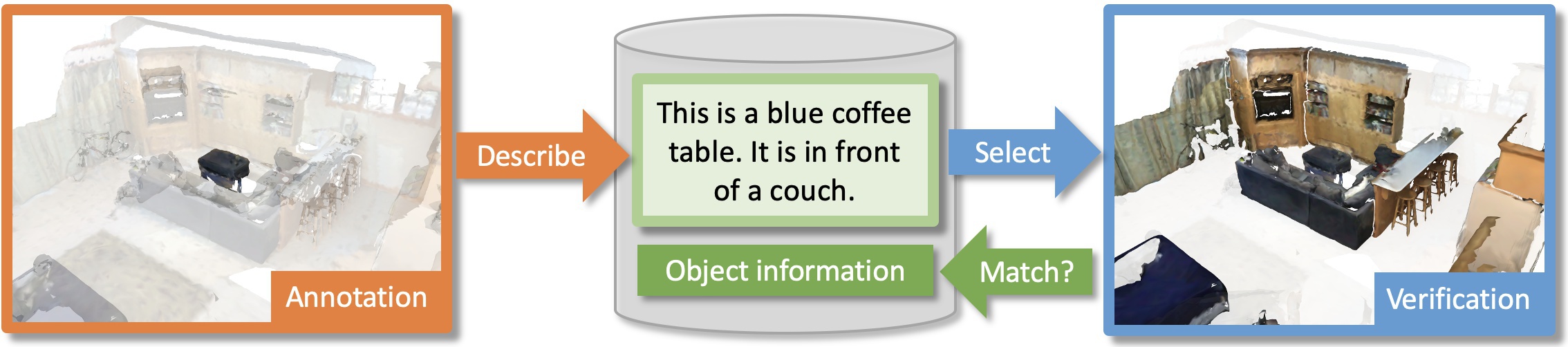}
    \caption{Our data collection pipeline. The annotator writes a description for the focused object in the scene. Then, a verifier selects the objects that match the description. The selected object is compared with the target object to check that it can be uniquely identified by the description.}
    \label{fig:collection}
\end{figure}

\mypara{3D Vision and Language.}
Vision and language research is gaining popularity in image domains (e.g., image captioning~\citep{karpathy2015deep, vinyals2015show, xu2015show, lu2017knowing}, image-text matching~\citep{feng2014cross, kiros2014unifying, li2017identity, huang2017instance, gu2018look}, and text-to-image generation~\citep{reed2016generative, gu2018look, sharma2018chatpainter}), but there is little work on vision and language in 3D.
Chen et al.~\citep{chen2018text2shape} learn a joint embedding of 3D shapes from ShapeNet~\citep{chang2015shapenet} and corresponding natural language descriptions.
Achlioptas et al.~\citep{achlioptas2019shapeglot} disambiguate between different objects using language.
Recent work has started to investigate grounding of language to 3D by identifying 3D bounding boxes of target objects for simple arrangements of primitive shapes of different colors~\citep{prabhudesai2019embodied}.
Instead of focusing on isolated objects, we consider large 3D RGB-D reconstructions that are typical in semantic 3D scene understanding.
\updated{A closely related work by Kong et al.~\citep{kong2014you} studied the problem of coreference in text description of single-view RGB-D images of scenes, where they aimed to connect noun phrases in a scene description to 3D bounding boxes of objects.} Concurrent with this work, Achlioptas et al.~\citep{achlioptasreferit3d} introduces a new dataset and task that focuses on disambiguating objects from the same category with known localizations.


\vspace{-0.3cm}
\section{Task}

We introduce the task of object localization in 3D scenes using natural language (Fig.~\ref{fig:overview}).
The input is a 3D scene and free-form text describing an object in the scene.
The scene is represented as a point cloud with additional features such as colors and normals for each point.
The goal is to predict the 3D bounding box of the object that matches the input description.


\section{Dataset}

The \DATASET~dataset is based on ScanNet~\citep{dai2017scannet} which is composed of \updated{1,613} RGB-D scans taken in \updated{806} unique indoor environments.
We provide $5$ descriptions for each object in each scene, focusing on complete coverage of all objects that are present in the reconstruction.
Here, we summarize the annotation process and statistics of our dataset (see supplement for more details).

\begin{figure}[t!]
    \begin{floatrow}
        \ffigbox{%
            \includegraphics[width=\linewidth]{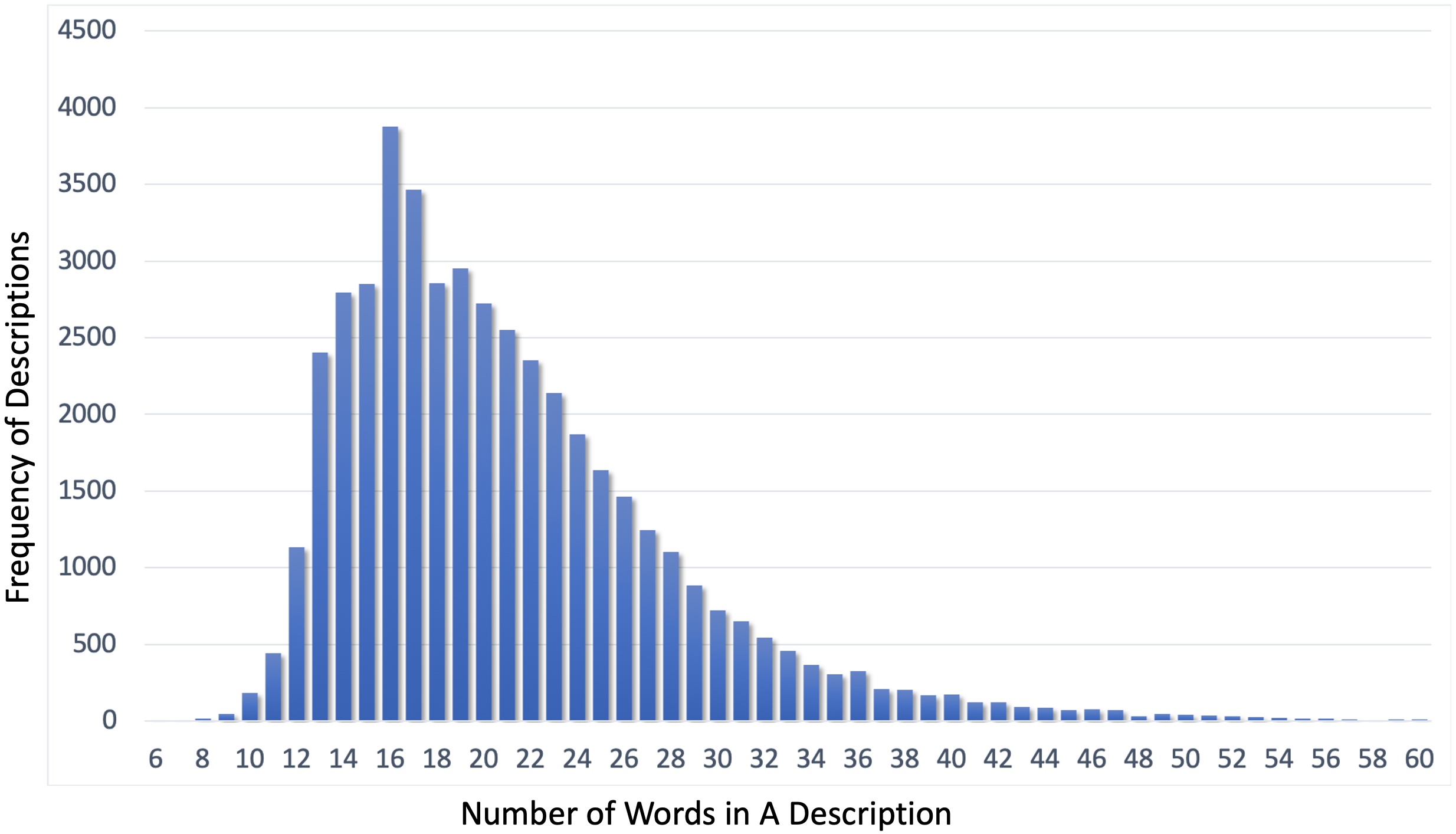}
        }{%
          \caption{Description lengths}
          \label{fig:length}%
        }
        \capbtabbox{%
            \begin{tabular}{l*{1}{c}}
                \toprule
                Number of descriptions & \text{\NUMDESCS} \\
                Number of scenes & \text{\NUMSCENES} \\
                Number of objects & \text{\NUMOBJECTS} \\
                Number of objects per scene & \text{\NUMOBJECTSPS} \\
                Number of descriptions per scene & \text{\NUMDESCSPS} \\
                Number of descriptions per object & \text{\AVGDESCS} \\
                Size of vocabulary & \text{\VOCABSIZE} \\
                Average length of descriptions & \text{\AVGLEN} \\
                \bottomrule
            \end{tabular}
        }{%
          \caption{\DATASET~dataset statistics.}
          \label{tab:dataset_stats}%
        }
    \end{floatrow}
\end{figure}

\begin{figure*}[t!]
    \centering
    \begin{subfigure}{0.19\textwidth}
        \centering
        \includegraphics[width=0.99\textwidth]{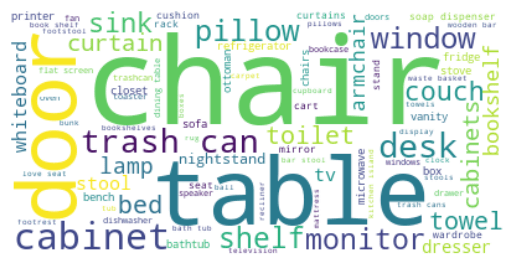}
        \caption{}
        \label{fig:wordcloud_object}
    \end{subfigure}
    \begin{subfigure}{0.19\textwidth}
        \centering
        \includegraphics[width=0.99\textwidth]{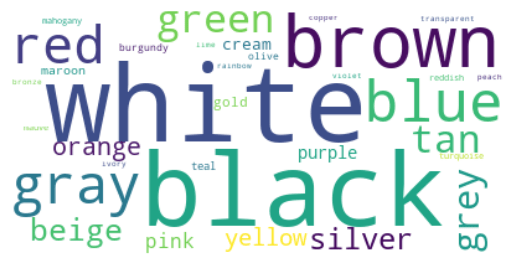}
        \caption{}
        \label{fig:wordcloud_color}
    \end{subfigure}
    \begin{subfigure}{0.19\textwidth}
        \centering
        \includegraphics[width=0.99\textwidth]{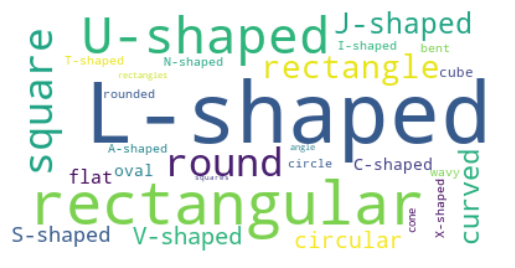}
        \caption{}
        \label{fig:wordcloud_shape}
    \end{subfigure}
    \begin{subfigure}{0.19\textwidth}
        \centering
        \includegraphics[width=0.99\textwidth]{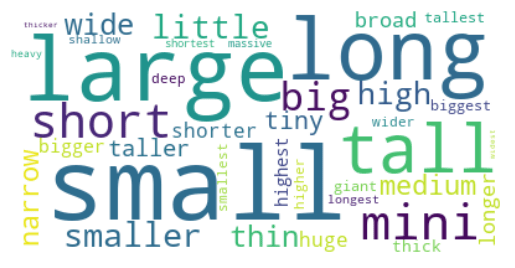}
        \caption{}
        \label{fig:wordcloud_size}
    \end{subfigure}
    \begin{subfigure}{0.19\textwidth}
        \centering
        \includegraphics[width=0.99\textwidth]{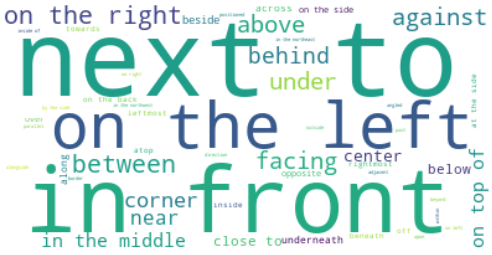}
        \caption{}
        \label{fig:wordcloud_spatial}
    \end{subfigure}
    \caption{Word clouds of terms for (a) object names (b) colors (c) shapes (d) sizes, and (e) spatial relations for the \DATASET~dataset.  Bigger fonts indicate more frequent terms in the descriptions.}
    \label{fig:wordcloud}
\end{figure*}

\subsection{Data Collection}
We deploy a web-based annotation interface on Amazon Mechanical Turk (AMT) to collect object descriptions in the ScanNet scenes.
The annotation pipeline consists of two stages: i) description collection, and ii) verification (Fig.~\ref{fig:collection}).
From each scene, we select objects to annotate by restricting to indoor furniture categories and excluding structural objects such as ``Floor'' and ``Wall''.
We manually check the selected objects are recognizable and filter out objects with reconstructions that are too incomplete or hard to identify.

\mypara{Annotation}
The 3D web-based UI shows each object in context.
The workers see all objects other than the target object faded out and a set of captured image frames to compensate for incomplete details in the reconstructions.
The initial viewpoint is random but includes the target object.
Camera controls allow for adjusting the camera view to better examine the target object. 
We ask the annotator to describe the appearance of the target and its spatial location relative to other objects.
To ensure the descriptions are informative, we require the annotator to provide at least two full sentences.
We batch and randomize the tasks so that each object is described by five different workers.

\mypara{Verification}
We recruit trained workers (students) to verify that the descriptions are discriminative and correct.
Verifiers are shown the 3D scene and a description, and are asked to select the objects (potentially multiple) in the scene that match the description.
Descriptions that result in the wrong object or multiple objects are filtered out.
Verifiers also correct spelling and wording issues in the description when necessary.
We filter out 2,823 invalid descriptions that do not match the target objects and fix writing issues for 2,129 descriptions.

\subsection{Dataset Statistics}
We collected \text{\NUMDESCS} descriptions for \text{\NUMSCENES} ScanNet scenes\footnote{\updated{6} scenes are excluded since they do not contain any objects to describe}.
On average, there are \text{\NUMOBJECTSPS} objects, \text{\NUMDESCSPS} descriptions per scene, and \text{\AVGDESCS} descriptions per object after filtering (see Tab.~\ref{tab:dataset_stats} for basic statistics, Tab.~\ref{tab:samples} for sample descriptions, and Fig.~\ref{fig:length} for the distribution of the description lengths).
The descriptions are complex and diverse, covering over 250 types of common indoor objects, and exhibiting interesting linguistic phenomena.
Due to the complexity of the descriptions, one of the key challenges of our task is to determine what parts of the description describe the target object, and what parts describe neighboring objects.
Among those descriptions, 41,034 mention object attributes such as color, shape, size, etc.
We find that many people use spatial language ($98.7\%$), color ($74.7\%$), and shape terms ($64.9\%$).
In contrast, only $14.2$\% of the descriptions convey size information.
Fig~\ref{fig:wordcloud} shows commonly used object names and attributes.
Tab.~\ref{tab:samples} shows interesting expressions, including comparatives (``taller'') and superlatives (``the biggest one''), as well as phrases involving ordinals such as ``third from the wall''.
Overall, there are 672 and 2,734 descriptions with comparative and superlative phrases.
We provide more detailed statistics in the supplement.



 \begin{table*}[t]
    \centering
    \resizebox{\columnwidth}{!}{
        \begin{tabular}{l}
            \toprule
            1. There is a brown wooden chair placed right against the wall. \\
            2. This is a triangular shape table. The table is near the armchair. \\
            3. The little nightstand. The nightstand is on the right of the bed. \\
            4. This is a short trash can. It is in front of a taller trash can. \\
            5. The couch is the biggest one below the picture. The couch has three seats and is brown. \\
            6. This is a gray desk chair. This chair is the last one on the side closest to the open door. \\
            7. The kitchen counter is covering the lower cabinets. The kitchen counter is under the upper \\ 
            \hspace{4mm}cabinets that are mounted above. \\
            8. This is a round bar stool. It is third from the wall. \\
            \bottomrule
        \end{tabular}
    }
    \caption{Examples from our dataset illustrating different types of phrases such as attributes (1-8) and parts (5), comparatives (4), superlatives (5), intra-class spatial relations (6), inter-class spatial relations (7) and ordinal numbers (8). }
    \label{tab:samples}
\end{table*}

\section{Method}

\begin{figure*}[!t]
    \centering
    \includegraphics[width=\textwidth]{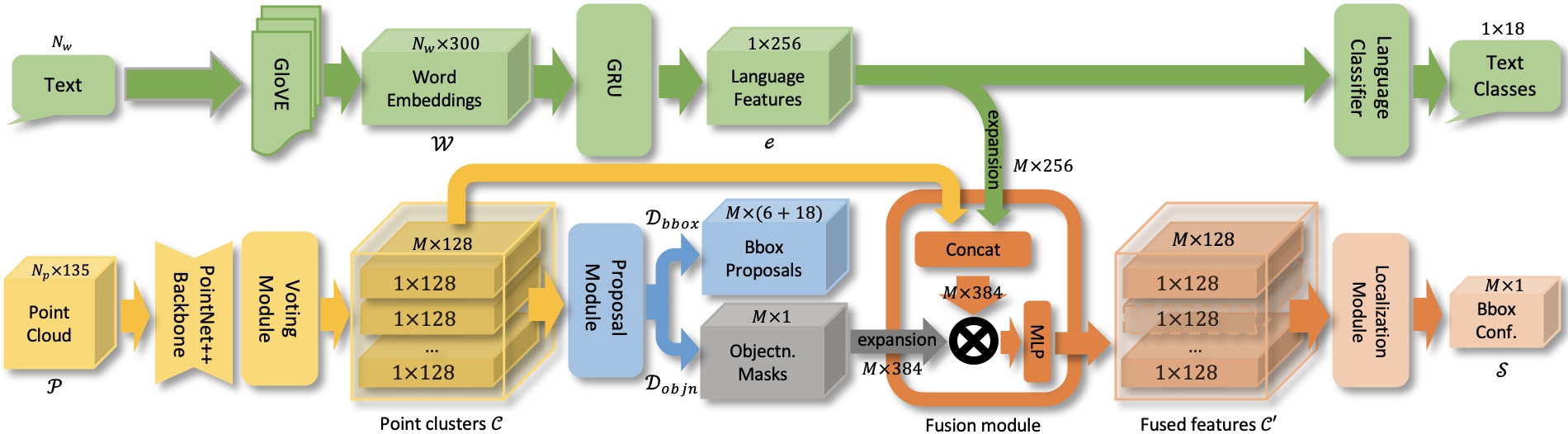}
    \caption{\METHOD~architecture:
    The PointNet++ \citep{qi2017pointnet++} backbone takes as input a point cloud and aggregates it to high-level point feature maps, which are then clustered and fused as object proposals by a voting module similar to Qi et al.~\citep{qi2019deep}. Object proposals are masked by the objectness predictions, and then fused with the sentence embedding of the input descriptions, which is obtained by a GloVE \citep{pennington2014glove} + GRU \citep{chung2014empirical} embedding. In addition, an extra language-to-object classifier serves as a proxy loss. We apply a softmax function in the localization module to output the confidence scores for the object proposals.}
    \label{fig:architecture}
\end{figure*}

Our architecture consists of two main modules: 1) detection \& encoding; 2) fusion \& localization (Fig.~\ref{fig:architecture}).
The detection \& encoding module encodes the input point cloud and description, and outputs the object proposals and the language embedding, which are fed into the fusion module to mask out invalid object proposals and produce the fused features.
Finally, the object proposal with the highest confidence predicted by the localization module is chosen as the final output.

\subsection{Data Representations}
\mypara{Point clouds} We randomly sample $N_P$ vertices of one scan from ScanNet as the input point cloud $\mathcal{P}=\{(p_{i}, f_{i})\}$, where $p_i \in \mathcal{R}^{3}$ represents the point coordinates in 3D space and $f_i$ stands for additional point features such as colors and normals. Note that the point coordinates $p_i$ provides only geometrical information and does not contain other visual information such as color and texture. Since descriptions of objects do refer to attributes such as color and texture, we incorporate visual appearance by adapting the feature projection scheme in Dai et al.~\citep{dai20183dmv} to project multi-view image features $v_i \in \mathcal{R}^{128}$ to the point cloud. The image features are extracted using a pre-trained ENet~\citep{paszke2016enet}.  Following Qi et al.~\citep{qi2019deep}, we also append the height of the point from the ground and normals to the new point features $f_i' \in \mathcal{R}^{135}$. The final point cloud data is prepared offline as $\mathcal{P}'=\{(p_{i}, f_i')\} \in \mathcal{R}^{N_P \times 135}$. We set $N_P$ to $40,000$ in our experiments. 

\mypara{Descriptions} We tokenize the input description with SpaCy~\citep{spacy2}
and the $N_W$ tokens to $300$-dimensional word embedding vectors $\mathcal{W}=\{w_{j}\} \in \mathcal{R}^{N_W \times 300}$ using pretrained GloVE word embeddings~\citep{pennington2014glove}.

\subsection{Network Architecture}
Our method takes as input the preprocessed point cloud $\mathcal{P}'$ and the word embedding sequence $\mathcal{W}$ representing the input description and outputs the 3D bounding box for the proposal which is most likely referred to by the input description. Conceptually, our localization pipeline consists of the following four stages: detection, encoding, fusion and localization.

\mypara{Detection} 
As the first step in our network, we detect all probable objects in the given point cloud.
To construct our detection module, we adapt the PointNet++~\citep{qi2017pointnet++} backbone and the voting module in Qi et al.~\citep{qi2019deep} to process the point cloud input and aggregate all object candidates to individual clusters. The output from the voting module is a set of point clusters $\mathcal{C} \in \mathcal{R}^{M \times 128}$ representing all object proposals with enriched point features, where $M$ is the upper bound of the number of proposals. Next, the proposal module takes in the point clusters and processes those clusters to predict the objectness mask $\mathcal{D}_{\text{objn}} \in \mathcal{R}^{M \times 1}$ and the axis-aligned bounding boxes $\mathcal{D}_{\text{bbox}} \in \mathcal{R}^{M \times (6+18)}$ for all $M$ proposals, where each $\mathcal{D}_{\text{bbox}}^{i} = (c_x, c_y, c_z, r_x, r_y, r_z, l)$ consists of the box center $c$, the box lengths $r$ and a vector $l \in \mathcal{R}^{18}$ representing the semantic predictions.

\mypara{Encoding} The sequences of word embedding vectors of the input description are fed into a GRU cell \citep{chung2014empirical} to aggregate the textual information. We take the final hidden state $e \in \mathcal{R}^{256}$ of the GRU cell as the final language embedding.

\mypara{Fusion} The outputs from the previous detection and encoding modules are fed into the fusion module (orange block in Fig.~\ref{fig:architecture}, see supplemental for details) to integrate the point features together with the language embeddings. Specifically, each feature vector $c_i \in \mathcal{R}^{128}$ in the point cluster $\mathcal{C}$ is concatenated with the language embedding $e \in \mathcal{R}^{256}$ as the extended feature vector, which is then masked by the predicted objectness mask $\mathcal{D}_{\text{objn}}^i \in \{0, 1\}$ and fused by a multi-layer perceptron as the the final fused cluster features $C' = \{c_i'\} \in \mathcal{R}^{M \times 128}$.

\mypara{Localization} The localization module aims to predict which of the proposed bounding boxes corresponds to the description.  Point clusters with fused cluster features $\mathcal{C}'=\{c_i'\}$ are processed by a single layer perceptron to produce the raw scores of how likely each box is the target box. We use a softmax function to squash all the raw scores into the interval of $[0,1]$ as the localization confidences $S=\{s_i\} \in \mathcal{R}^{M \times 1}$ for the proposed $M$ bounding boxes.

\subsection{Loss Function}
\mypara{Localization loss}
For the predicted localization confidence $s_i \in [0, 1]$ for object proposal $\mathcal{D}_{\text{bbox}}^i$, the target label is represented as $t_i \in \{ 0, 1 \}$. Following the strategy of Yang et al.~\citep{yang2019fast}, we set the label $t_j$ for the $j^{th}$ box that has the highest IoU score with the ground truth box as $1$ and others as $0$. \updated{We then use a cross-entropy loss as the localization loss $\mathcal{L}_{\text{loc}} = -\sum_{i=1}^{M}t_i\log(s_i)$.
}\\
\mypara{Object detection loss}
We use the same detection loss $\mathcal{L}_{det}$ as introduced in Qi et al.~\citep{qi2019deep} for object proposals $\mathcal{D}_{\text{bbox}}^i$ and $\mathcal{D}_{\text{objn}}^i$: $\mathcal{L}_{\text{det}} = \mathcal{L}_{\text{vote-reg}} + 0.5\mathcal{L}_{\text{objn-cls}} + \mathcal{L}_{\text{box}} + 0.1\mathcal{L}_{\text{sem-cls}}$, where $\mathcal{L}_{\text{vote-reg}}$, $\mathcal{L}_{\text{objn-cls}}$, $\mathcal{L}_{\text{box}}$ and $\mathcal{L}_{\text{sem-cls}}$ represent the vote regression loss (defined in Qi et al.~\citep{qi2019deep}), the objectness binary classification loss, box regression loss and the semantic classification loss for the $18$ ScanNet benchmark classes, respectively. We ignore the bounding box orientations in our task and simplify $\mathcal{L}_{\text{box}}$ as $\mathcal{L}_{\text{box}} = \mathcal{L}_{\text{center-reg}} + 0.1\mathcal{L}_{\text{size-cls}} + \mathcal{L}_{\text{size-reg}}$, where $\mathcal{L}_{\text{center-reg}}$, $\mathcal{L}_{\text{size-cls}}$ and $\mathcal{L}_{\text{size-reg}}$ are used for regressing the box center, classifying the box size and regressing the box size, respectively. We refer readers to Qi et al.~\citep{qi2019deep} for more details. 

\mypara{Language to object classification loss}
To further supervise the training, we include an object classification loss based on the input description. We consider the 18 ScanNet benchmark classes (excluding the label ``Floor'' and ``Wall''). The language to object classification loss $\mathcal{L}_{\text{cls}}$ is a multi-class cross-entropy loss.

\mypara{Final loss}
The final loss is a linear combination of the localization loss, object detection loss and the language to object classification loss: $\mathcal{L} = \alpha\mathcal{L}_{\text{loc}} + \beta\mathcal{L}_{\text{det}} + \gamma\mathcal{L}_{\text{cls}}$, where $\alpha$, $\beta$ and $\gamma$ are the weights for the individual loss terms. After fine-tuning on the validation split, we set those weights to $0.1$, $10$, and $1$ in our experiments to ensure the loss terms are roughly of the same magnitude.

\subsection{Training and Inference}
\mypara{Training}
During training, the detection and encoding modules propose object candidates as point clusters, which are then fed into the fusion and localization modules to fuse the features from the previous module and predict the final bounding boxes. We train the detection backbone end-to-end with the detection loss. In the localization module, we use a softmax function to compress the raw scores to $[0,1]$. The higher the predicted confidence is, the more likely the proposal will be chosen as output. To filter out invalid object proposals, we use the predicted objectness mask to ensure that only positive proposals are taken into account. We set the maximum number of proposals $M$ to $256$ in practice.

\mypara{Inference}
Since there can be overlapping detections, we apply a non-maximum suppression module to suppress those overlapping proposals in the inference step. The remaining object proposals are fed into the localization module to predict the final score for each proposal. The number of object proposals is less than the upper bound $M$ in the training step.  

\mypara{Implementation Details}
We implement our architecture using PyTorch and train the model end-to-end using ADAM~\citep{kingma2014adam} with a learning rate of $1$e$-3$.
We train the model for roughly $130,000$ iterations until convergence.
To avoid overfitting, we set the weight decay factor to $1$e$-5$ and apply data augmentations to our training data. For point clouds, we apply rotation about all three axes by a random angle in $[-\ang{5},\ang{5}]$ and randomly translate the point cloud within $0.5$ meters in all directions.
\updated{We rotate around all axes (not just up), since the ground alignment in ScanNet is imperfect.}


\section{Experiments}

\mypara{Train/Val/Test Split.} \updated{Following the official ScanNet \citep{dai2017scannet} split, we split our data into train/val/test sets with \text{\NUMTRAIN}, \text{\NUMVAL} and \text{\NUMTEST} samples respectively, ensuring disjoint scenes for each split.  Results and analysis are conducted on the val split (except for results in Tab. \ref{tab:comp_baseline} bottom). The test set is hidden and will be reserved for the {\DATASET} benchmark.}

\mypara{Metric.} To evaluate the performance of our method, we measure the thresholded accuracy where the positive predictions have higher intersection over union (IoU) with the ground truths than the thresholds. Similar to work with 2D images, we use Acc@$k$IoU as our metric, where the threshold value $k$ for IoU is set to $0.25$ and $0.5$ in our experiments. 

\begin{figure*}[!t]
    \setlength{\belowcaptionskip}{-4mm}
    \centering
    \includegraphics[width=0.9\textwidth]{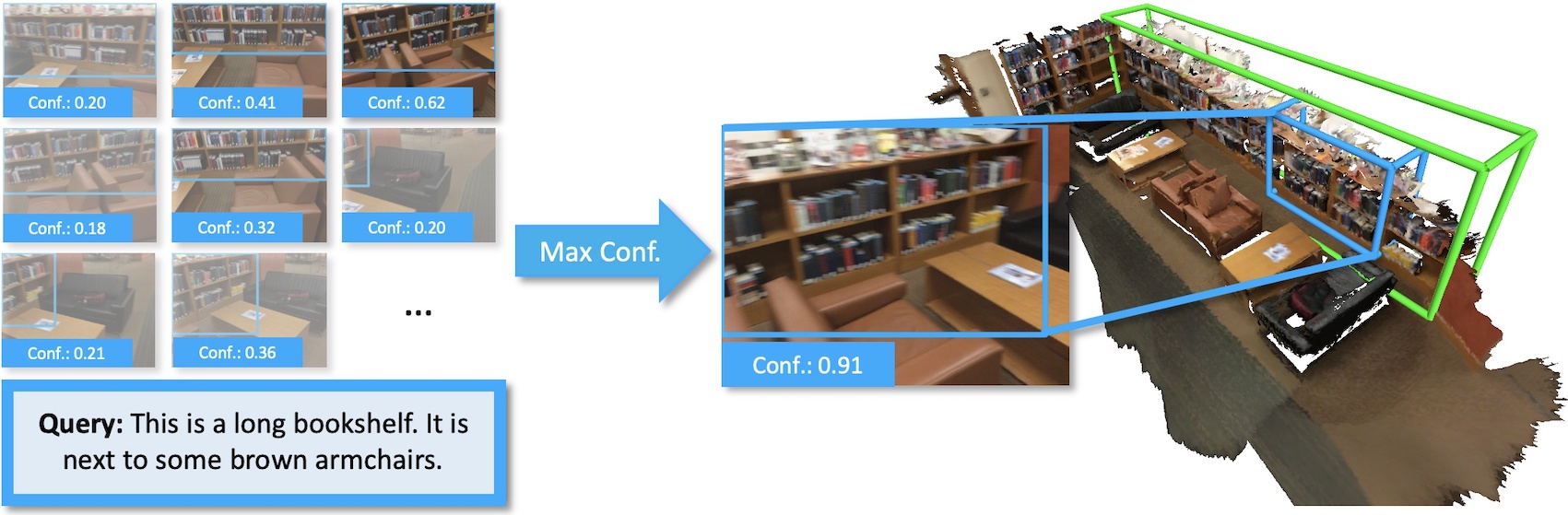}
    \caption{Object localization in an image using a 2D grounding  method and back-projecting the result to the 3D scene (\blue{blue} box) vs. directly localizing in the 3D scene (\green{green} box). Grounding in 2D images suffers from the limited view of a single frame, which results in inaccurate 3D bounding boxes.
    }
    \label{fig:2dproj}
\end{figure*}

\mypara{Baselines.}
We design several baselines by \updated{1) evaluating our language localization module on ground truth bounding boxes, 2) adapting 3D object detectors, and 3) adapting 2D referring methods to 3D using back-projection.} 

\mypara{\emph{OracleCatRand \& OracleRefer:} }
To examine the difficulty of our task, we use an oracle with ground truth bounding boxes of objects, and predict the box by simply selecting a random box that matches the object category (OracleCatRand) or our trained fusion and localization modules (OracleRefer).

\mypara{\emph{VoteNetRand \updated{\& VoteNetBest}:} }
From the predicted object proposals of the VoteNet backbone~\citep{qi2019deep}, we \updated{select one of the bounding box proposals, either by selecting a box randomly with the correct semantic class label (VoteNetRand) or the best matching box given the ground truth (VoteNetBest).  VoteNetBest provides an upper bound on how well the object detection component works for our task, while VoteNetRand provides a measure of whether additional information beyond the semantic label is required.} 

\mypara{\emph{SCRC \& One-stage:}}
2D image baselines for referring expression comprehension by extending SCRC~\citep{hu2016natural} and One-stage~\citep{yang2019fast} to 3D using back-projection.
Since 2D referring expression methods operate on a single image frame, we construct a 2D training set by using the recorded camera pose associated with each annotation to retrieve the frame from the scan video with the closest camera pose.  At inference time, we sample frames from the scans (using every $20$th frame) and predict the target 2D bounding boxes in each frame.  We then select the 2D bounding box with the highest confidence score from the bounding box candidates and project it to 3D using the depth map for that frame (see Fig.~\ref{fig:2dproj}). 

\mypara{\emph{Ours:}}
We compare our full end-to-end model against using a pretrained VoteNet backbone with a trained GRU~\citep{chung2014empirical} for selecting a matching bounding box.

\subsection{Task Difficulty}
To understand how informative the input description is beyond capturing the object category, we analyze the performance of the methods on ``unique'' and ``multiple'' subsets with \text{\NUMUNIQUE} and \text{\NUMMULTIPLE} samples \updated{from val split}, respectively. The ``unique'' subset contains samples where only one unique object from a certain category matches the description, while the ``multiple'' subset contains ambiguous cases where there are multiple objects of the same category.
For instance, if there is only one refrigerator in a scene, it is sufficient to identify that the sentence refers to a refrigerator.  In contrast, if there are multiple objects of the same category in a scene (e.g., chair), the full description must be taken into account.  From the OracleCatRand baseline, we see that information from the description, other than the object category, is necessary to disambiguate between multiple objects (see Tab.~\ref{tab:comp_baseline} Acc@0.5IoU multiple).  From the OracleRefer baseline, we see that using our fused language module, we are able to improve beyond over selecting a random object of the same category (multiple Acc@0.5IoU increases from $17.84\%$ to \updated{$32.00\%$}), but we often fail to identify the correct object category (unique Acc@0.5IoU drops from $100.0\%$ to \updated{$73.55\%$}).  

\begin{table*}[!t]
    \centering
    \resizebox{\columnwidth}{!}{
        \begin{tabular}{l*{6}{c}}
            \toprule
            & \multicolumn{2}{c}{unique} & \multicolumn{2}{c}{multiple} & \multicolumn{2}{c}{overall} \\
            & Acc@0.25 & Acc@0.5 & Acc@0.25 & Acc@0.5 & Acc@0.25 & Acc@0.5 \\
            \midrule
            OracleCatRand (GT boxes + RandCat) & 100.00 & 100.00 & 18.09 & 17.84 & 29.99 & 29.76 \\
            OracleRefer (GT boxes + GRU) & \updated{74.09} & \updated{73.55} & \updated{32.57} & \updated{32.00}  & \updated{40.63} & \updated{40.06} \\
            \midrule
            \updated{VoteNetRand (VoteNet\citep{qi2019deep} + RandCat)} & \updated{34.34} & \updated{19.35} & \updated{5.73} & \updated{2.81} & \updated{10.00} & \updated{5.28} \\
            \updated{VoteNetBest (VoteNet\citep{qi2019deep} + Best)} & \updated{88.85} & \updated{85.50} & \updated{46.63} & \updated{46.42} & \updated{55.10} & \updated{54.33} \\
            \midrule
            SCRC \citep{hu2016natural} + backproj & 24.03 & 9.22 & 17.77 & 5.97 & 18.70 & 6.45 \\
            One-stage \citep{yang2019fast} + backproj & 29.32 & 22.82 & 18.72 & 6.49 & 20.38 & 9.04 \\
            \midrule
            \updated{Ours (VoteNet[48] + GRU)} & \updated{\textbf{77.33}} & \updated{51.73} & \updated{30.43} & \updated{19.46} & \updated{39.52} & \updated{25.72} \\
            \updated{Ours (end-to-end)} & 76.33 & \textbf{53.51} & \textbf{32.73} & \textbf{21.11} & \textbf{41.19} & \textbf{27.40} \\
 
            \midrule
            \multicolumn{7}{c}{Test results (\text{\DATASET} benchmark)} \\
            \midrule
            OracleRefer (GT boxes + GRU) & \updated{72.37} & \updated{71.84} & \updated{31.81} & \updated{31.26}  & \updated{39.69} & \updated{39.13} \\
            \updated{VoteNetBest (VoteNet\citep{qi2019deep} + Best)} & \updated{86.78} & \updated{83.85} & \updated{45.54} & \updated{45.33} & \updated{53.82} & \updated{53.07} \\
            \midrule
            \updated{Ours (VoteNet[48] + GRU)} & \updated{\textbf{72.55}} & \updated{\textbf{47.24}} & \updated{32.90} & \updated{19.16} & \updated{41.79} & \updated{25.45} \\
            \updated{Ours (end-to-end)}  & \updated{71.06} & \updated{46.66} & \updated{\textbf{35.17}} & \updated{\textbf{20.92}} & \updated{\textbf{43.22}} & \updated{\textbf{26.69}} \\
            \bottomrule
        \end{tabular}
    }
    \caption{Comparison of localization results obtained by our \text{\METHOD} and baseline models. We measure percentage of predictions whose IoU with the ground truth boxes are greater than 0.25 and 0.5. We also report scores on ``unique'' and ``multiple'' subsets; unique means that there is only a single object of its class in the scene. We outperform all baselines by a significant margin.}
    \label{tab:comp_baseline}
\end{table*}

\begin{figure*}[!t]
    \centering
    \includegraphics[width=0.9\textwidth]{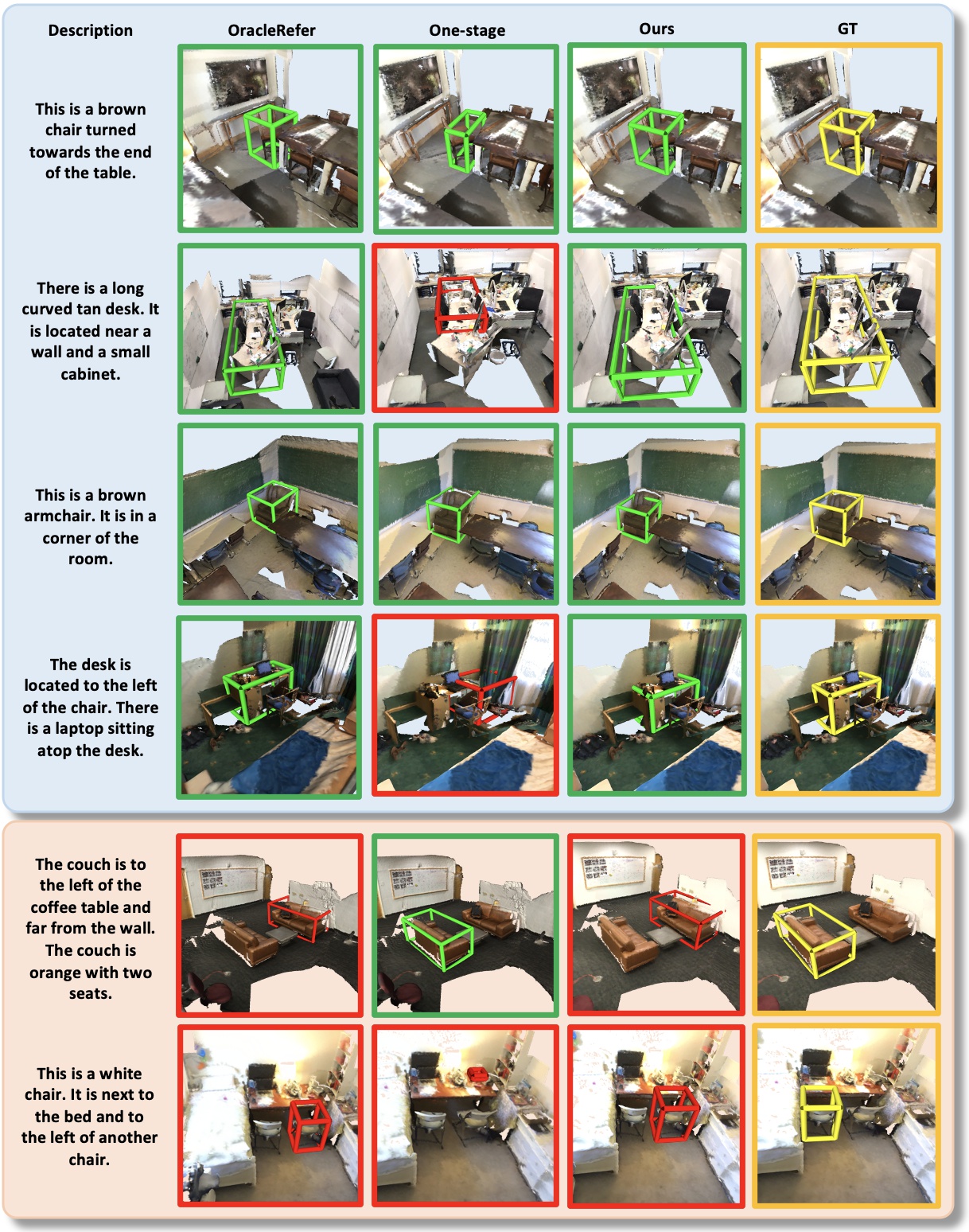}
    \caption{Qualitative results from baseline methods and \text{\METHOD}. Predicted boxes are marked \green{green} if they have an IoU score higher than 0.5, otherwise they are marked \red{red}. We show examples where our method produced good predictions (\blue{blue} block) as well as failure cases (\orange{orange} block). Image best viewed in color.}
    \label{fig:qualitative}
\end{figure*}

\subsection{Quantitative Analysis}
\updated{We evaluate the performance of our model against baselines on the val and the hidden test split of \text{\DATASET} which serves as the \text{\DATASET} benchmark (see Tab. \ref{tab:comp_baseline}).} 
\updated{Note that for all results using Ours and VoteNet for object proposal, we take the average of $5$ differently seeded subsamplings (of seed points and vote points) during inference (see supplemental for more details on experimental variance).}
\updated{Training the detection backbone jointly with the localization module (end-to-end) leads to a better performance when compared to the model trained separately (VoteNet\citep{qi2019deep} + GRU). However, as the accuracy gap between VoteNetBest and ours (end-to-end) indicates, there is still room for improving the match between language inputs and the visual signals.} \updated{For the val split, we also include additional experiments on the 2D baselines and a comparison with VoteNetRand.}
With just category information, VoteNetRand is able to perform relatively well on the ``unique'' subset, but has trouble identifying the correct object in the ``multiple'' case.  However, the gap between the VoteNetRand and OracleCatRand for the ``unique'' case shows that 3D object detection still need to be improved. Our method is able to improve over the bounding box predictions from VoteNetRand, and leverages additional information in the description to differentiate between ambiguous objects. It adapts better to the 3D context compared to the 2D methods (SCRC and One-stage) which is limited by the view of a single frame (see Fig.~\ref{fig:2dproj} and Fig.~\ref{fig:qualitative}).

\subsection{Qualitative Analysis}
Fig.~\ref{fig:qualitative} shows results produced by OracleRefer, One-stage, and our method. The successful localization cases in the green boxes show our architecture can handle the semantic correlation between the scene contexts and the textual descriptions. In contrast, even provided with a pool of ground truth proposals, OracleRefer sometimes still fails to predict correct bounding boxes, while One-stage is limited by the single view and hence cannot produce accurate bounding boxes in 3D space. The failure case of OracleRefer suggests that our fusion \& localization module can still be improved. Some failure cases of our method are displayed in the orange block in Fig.~\ref{fig:qualitative}, indicating that our architecture cannot handle all spatial relations to distinguish between ambiguous objects.

\subsection{Ablation Studies}

\begin{table*}[!t]
    \centering
    \resizebox{\columnwidth}{!}{
        \begin{tabular}{l*{6}{c}}
            \toprule
            & \multicolumn{2}{c}{unique} & \multicolumn{2}{c}{multiple} & \multicolumn{2}{c}{overall} \\
            & Acc@0.25 & Acc@0.5 & Acc@0.25 & Acc@0.5 & Acc@0.25 & Acc@0.5 \\
            \midrule
            \updated{Ours (xyz)} & 63.98 & 43.57 & 29.28 & 18.99 & 36.01 & 23.76 \\					
            \updated{Ours (xyz+rgb)} & 63.24 & 41.78 & 30.06 & 19.23 & 36.50 & 23.61 \\					
            \updated{Ours (xyz+rgb+normals)} & 64.63 & 43.65 & 31.89 & 20.77 & 38.24 & 25.21 \\
            \updated{Ours (xyz+multiview)} & 77.20 & 52.69 & 32.08 & 19.86 & 40.84 & 26.23 \\
            \updated{Ours (xyz+multiview+normals)} & \textbf{78.22} & 52.38 & 33.61 & 20.77 & \textbf{42.27} & 26.90 \\
            \midrule
            \updated{Ours (xyz+lobjcls)} & 64.31 & 44.04 & 30.77 & 19.44 & 37.28 & 24.22 \\ 
            \updated{Ours (xyz+rgb+lobjcls)} & 65.00 & 43.31 & 30.63 & 19.75 & 37.30 & 24.32 \\ 
            \updated{Ours (xyz+rgb+normals+lobjcls)} & 67.64 & 46.19 & 32.06 & \textbf{21.26} & 38.97 & 26.10 \\
            \updated{Ours (xyz+multiview+lobjcls)} & 76.00 & 50.40 & \textbf{34.05} & 20.73 & 42.19 & 26.50 \\
            \updated{Ours (xyz+multiview+normals+lobjcls)} & 76.33 & \textbf{53.51} & 32.73 & 21.11 & 41.19 & \textbf{27.40} \\
            \bottomrule
        \end{tabular}
    }
    \caption{Ablation study with different features. We measure the percentages of predictions whose IoU with the ground truth boxes are greater than 0.25 and 0.5. Unique means that there is only a single object of its class in the scene.}
    \label{tab:ablation}
\end{table*}

We conduct an ablation study on our model to examine what components and point cloud features contribute to the performance (see Tab.~\ref{tab:ablation}).

\mypara{Does a language-based object classifier help?}
To show the effectiveness of the extra supervision on input descriptions, we conduct an experiment with the language to object classifier (+lobjcls) and without. Architectures with a language to object classifier outperform ones without it.  This indicates that it is helpful to predict the category of the target object based on the input description.

\mypara{Do colors help?}
We compare our method trained with the geometry and multi-view image features (xyz+multiview+lobjcls) with a model trained with only geometry (xyz+lobjcls) and one trained with RGB values from the reconstructed meshes (xyz+rgb+lobjcls). \text{\METHOD} trained with geometry and pre-processed multi-view image features outperforms the other two models. The performance of models with color information are higher than those that use only geometry.

\mypara{Do other features help?}
We include normals from the ScanNet meshes to the input point cloud features and compare performance against networks trained without them. The additional 3D information improves performance. Our architecture trained with geometry, multi-view features, and normals (xyz+multiview+ normals+lobjcls) achieves the best performance among all ablations.


\section{Conclusion}
In this work, we introduce the task of localizing a target object in a 3D point cloud using natural language descriptions. We collect the \DATASET dataset which contains \text{\NUMDESCS} unique descriptions for \text{\NUMOBJECTS} objects from \text{\NUMSCENES} ScanNet \citep{dai2017scannet} scenes. We propose an end-to-end method for localizing an object with a free-formed description as reference, which first proposes point clusters of interest and then matches them to the embeddings of the input sentence. Our architecture is capable of learning the semantic similarities of the given contexts and regressing the bounding boxes for the target objects. 
Overall, we hope that our new dataset and method will enable future research in the 3D visual language field.


\section*{Acknowledgements}
We would like to thank the expert annotators Josefina Manieu Seguel and Rinu Shaji Mariam, all anonymous workers on Amazon Mechanical Turk and the student volunteers (Akshit Sharma, Yue Ruan, Ali Gholami, Yasaman Etesam, Leon Kochiev, Sonia Raychaudhuri) at Simon Fraser University for their efforts in building the \text{\DATASET} dataset, and Akshit Sharma for helping with statistics and figures.
This work is funded by Google (AugmentedPerception), the ERC Starting Grant Scan2CAD (804724), and a Google Faculty Award. We would also like to thank the support of the TUM-IAS Rudolf M{\"o}{\ss}bauer and Hans Fischer Fellowships (Focus Group Visual Computing), as well as the the German Research Foundation (DFG) under the Grant \textit{Making Machine Learning on Static and Dynamic 3D Data Practical}.
Angel X. Chang is supported by the Canada CIFAR AI Chair program.
Finally, we thank Angela Dai for the video voice-over.

%
%
\bibliographystyle{splncs04}
\bibliography{egbib}

\newpage
\appendix
\section*{Supplementary Material}

\begin{figure}[!htb]
    \centering
    \includegraphics[width=\linewidth]{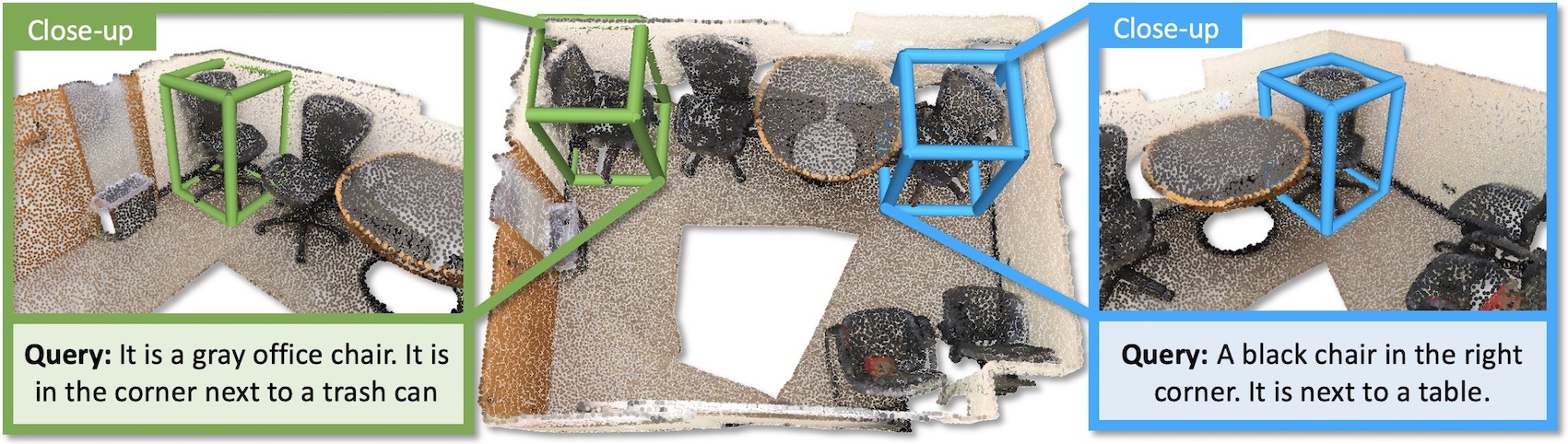}
    \caption{
    ScanRefer localizes objects in a scene given a language description as input. In many cases, including this example, there are multiple objects from the same category in a single scene which makes the problem challenging and interesting at the same time.
    }
    \label{fig:pointcloud_teaser}
\end{figure}

In this supplementary material, we provide addition details on the data collection and statistic of the ScanRefer dataset  (Section \ref{sec:supdataset}); we also provide implementation details of our localization network (Section \ref{sec:implemention}), as well as additional quantitative (Section \ref{sec:supquant}) and qualitative comparisons (Section \ref{sec:supqual}).

\section{Dataset}
\label{sec:supdataset}

\subsection{Statistics}
\label{sec:statistics}

\begin{figure}[!htb]
    \centering
    \includegraphics[width=\linewidth]{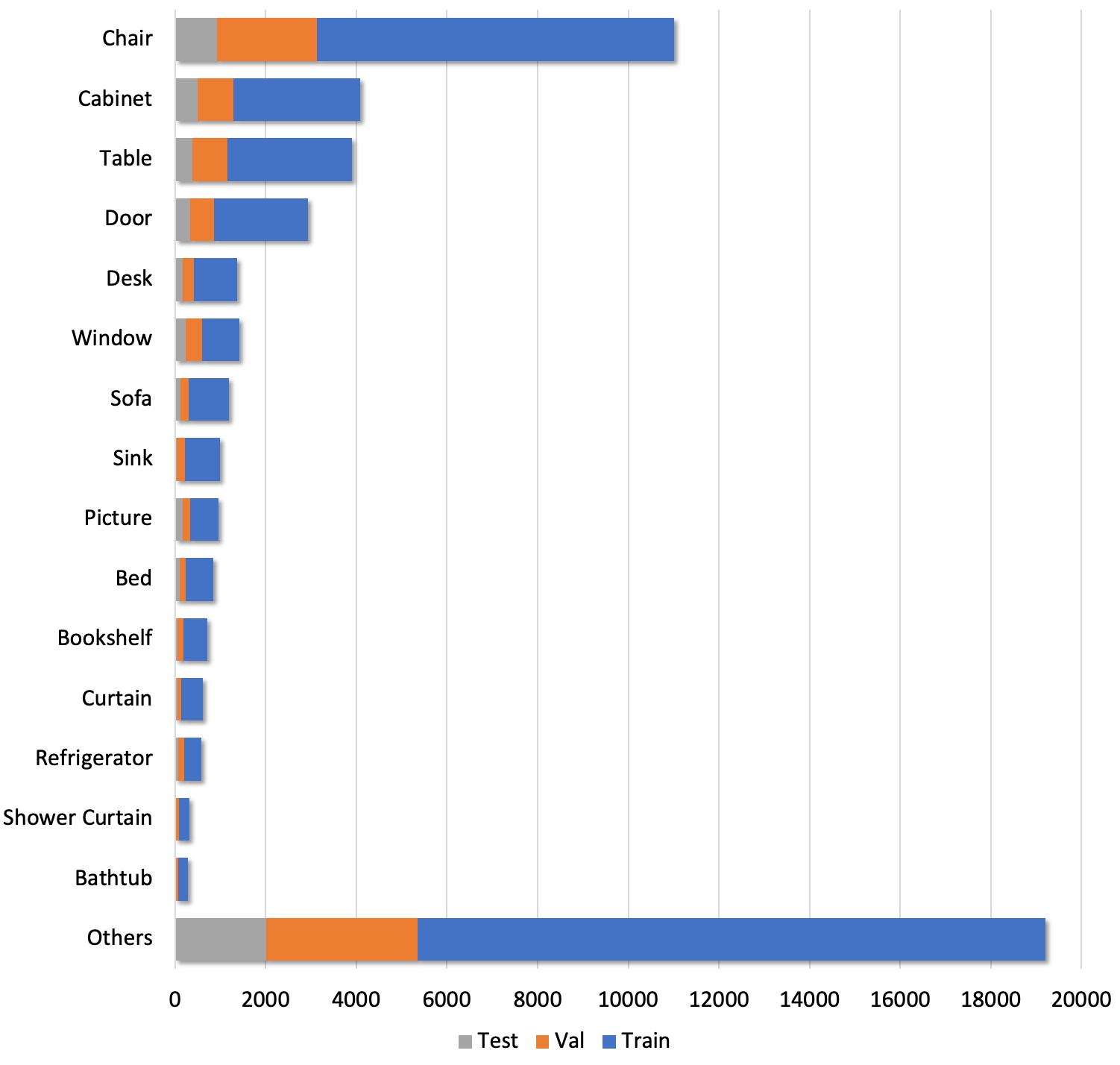}
    \caption{Distribution of categories of objects in the \text{\DATASET} dataset with annotated language descriptions.}
    \label{fig:distribution}
\end{figure}

We present the distribution of categories of the \text{\DATASET} dataset in Fig.~\ref{fig:distribution}. \text{\DATASET} provides a large coverage of furniture (e.g., chair, table, cabinet, bed, etc.) in indoor environments with various sizes, colors, materials, and locations. We use the same category names as in the original ScanNet dataset~\citep{dai2017scannet}. In total, we annotate \updated{11,046} objects from 265 categories from ScanNet~\citep{dai2017scannet}. Following the ScanNet voxel labeling task, we aggregate these finer-grained categories into 17 coarse categories and group the remaining object types into ``Others'' for a total of 18 object categories that we use to train the language-based object classifier. 

\begin{table*}[!t]
    \centering
    \begin{tabular}{l r r r r}
        \toprule
        & Train & Val & Test & Total \\
        \midrule
        Number of descriptions & \text{\NUMTRAIN} & \text{\NUMVAL} & \text{\NUMTEST} & \text{\NUMDESCS} \\
        Number of scenes & 562 & 141 & 97 & \text{\NUMSCENES} \\
        Number of objects & 7,875 & 2,068 & 1,103 & 11,046 \\
        Number of objects per scene & 14.01 & 14.67 & 11.37 & 14.14 \\
        Number of descriptions per scene & 65.24 & 67.43 & 55.77 & 65.68 \\
        Number of descriptions per object  & 4.66 & 4.60 & 4.90 & 4.64 \\
        \bottomrule
    \end{tabular}
    \caption{\text{\DATASET} dataset statistics on Train and Val splits.}
    \label{tab:dataset_split}
\end{table*}

\begin{table*}[!t]
    \centering
    \begin{tabular}{r c c c}
        \toprule
        Number of objects per scene & Unique & Multiple & Overall \\
        \midrule
        total & 3.00 & 11.81 & 14.14 \\
        same category as the target object & 1.00 & 4.96 & 2.98 \\
        \bottomrule
    \end{tabular}
    \caption{Average number of objects (per scene) for the ``Unique'' and ``Multiple'' subsets of the \DATASET~dataset.  Assuming ground truth bounding boxes, there are on average $14$ different objects for to disambiguate between.  For the ``Multiple'' subset, there are on average $5$ objects to disambiguate between even if we could match the semantic class perfectly.}
    \label{tab:dataset_difficulty}
\end{table*}

\begin{figure}[!ht]
    \centering
    \includegraphics[width=\linewidth]{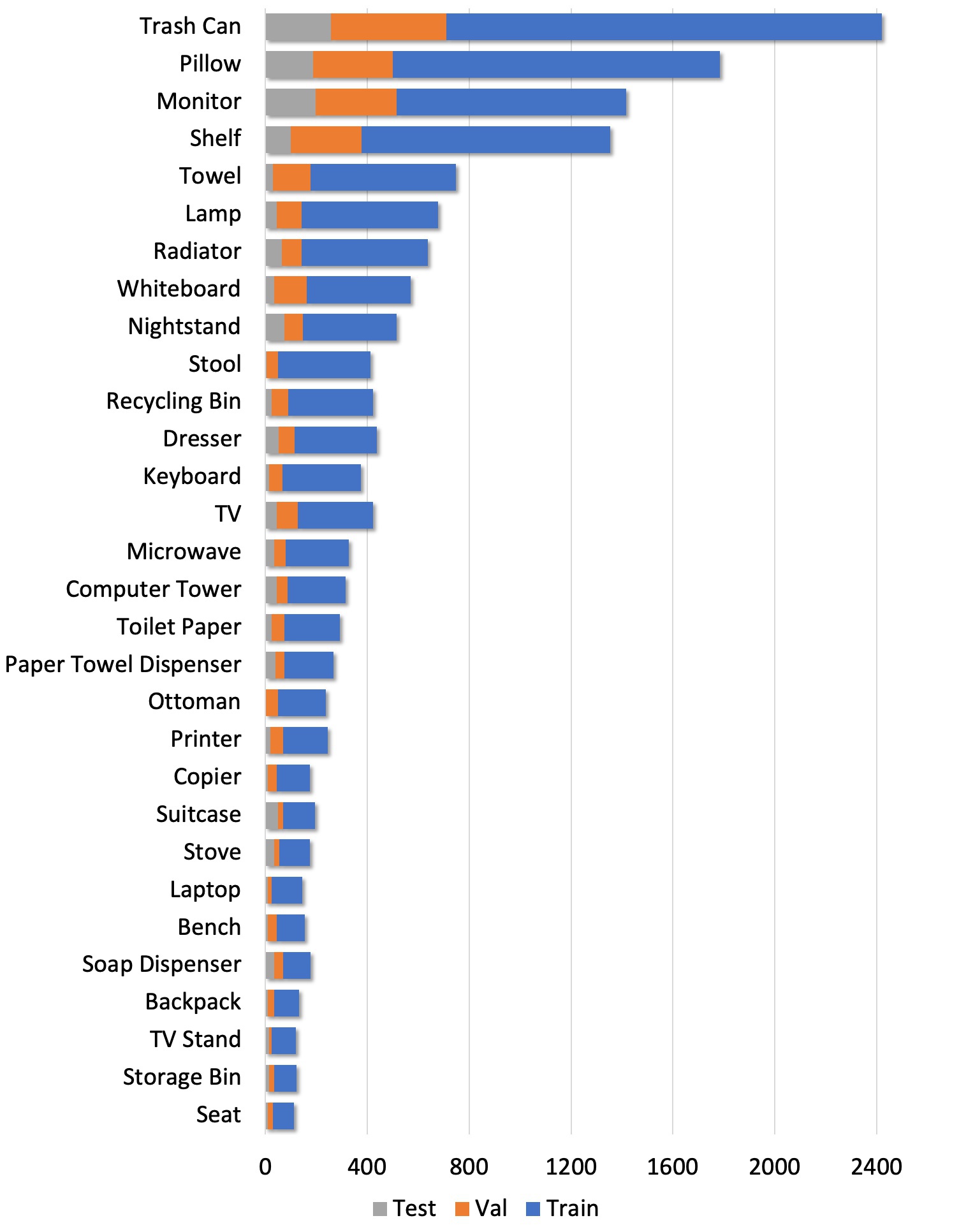}
    \caption{\updated{Distribution of the top 30 categories in the ``Others'' category of the Train/Val/Test splits of the \text{\DATASET} dataset (sorted in descending order according to the number of objects in the Train split).}}
    \label{fig:others}
\end{figure}

\begin{figure}[!ht]
    \centering
    \includegraphics[width=\linewidth]{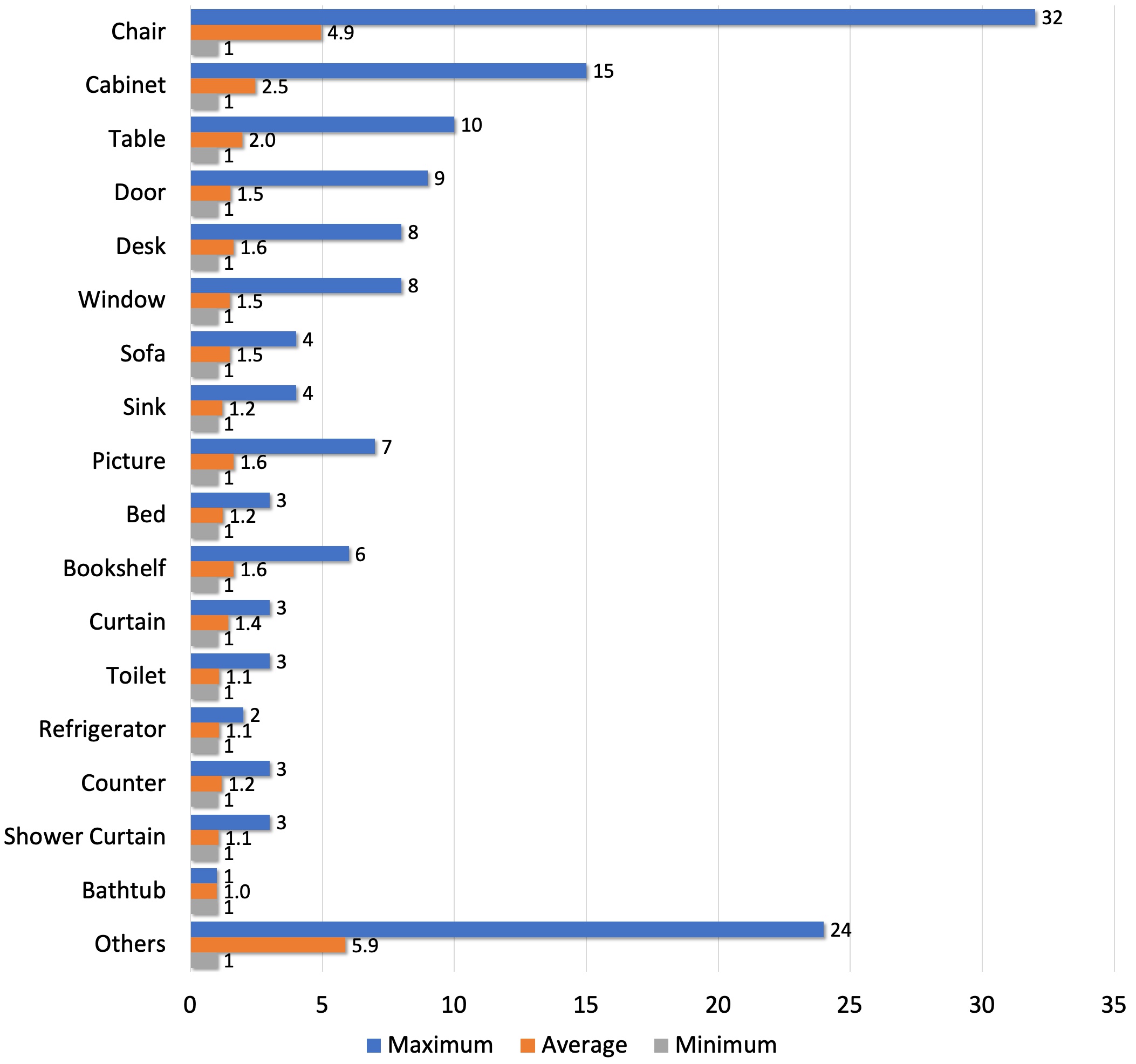}
    \caption{Average and maximum numbers of objects in each category per scene in the \text{\DATASET} dataset. For each category, we only consider scenes that contains the corresponding objects.}
    \label{fig:category_per_scene}
\end{figure}

Fig.~\ref{fig:others} shows the distribution of finer-grained objects in the category ``Others''. For each of the 18 coarse categories, Fig.~\ref{fig:category_per_scene} shows the average and maximum number of objects for that category in a scene in which an object of that category appears.  For instance, for scenes that contains a bed, the average number of beds is 1.22 and the maximum is 3.  

We also provide detailed statistics in our training and validation splits in Tab.~\ref{tab:dataset_split}. To further address the difficulty of our task, we present additional details about the ``unique'' and ``multiple'' subsets in Tab.~\ref{tab:dataset_difficulty}.  The ``unique'' subset consists of cases where there is just one unique object of that category (from the 18 ScanNet classes), in the scene.  In these cases, the object can be localized (assuming perfect object detection) just by identifying the semantic class of the target object from the description (e.g.,  localizing the table in the scene Fig.~\ref{fig:pointcloud_teaser}).  The ``multiple'' subset refers to cases where there are multiple objects of the same category as the target object in the scene, thus requiring disambiguation between multiple objects of the same time (e.g., localizing a specific chair in the scene in Fig.~\ref{fig:pointcloud_teaser}). As shown in Tab.~\ref{tab:dataset_difficulty}, since there are on average more objects of the same category as the target object in the ``multiple'' subset than in the ``unique'', it is more challenging to correctly localize the target object in the ``multiple'' subset.

\subsection{Collection Details}

In this section, we provide more details of the data annotation and verification processes of \text{\METHOD}.  %
The data collection took place over one month and involved 1,929 AMT workers. 
Together, the description collection and verification took around 4,984 man hours in total.

\begin{figure*}[t!]
    \centering
    \begin{subfigure}{0.9\textwidth}
        \centering
        \includegraphics[width=\textwidth]{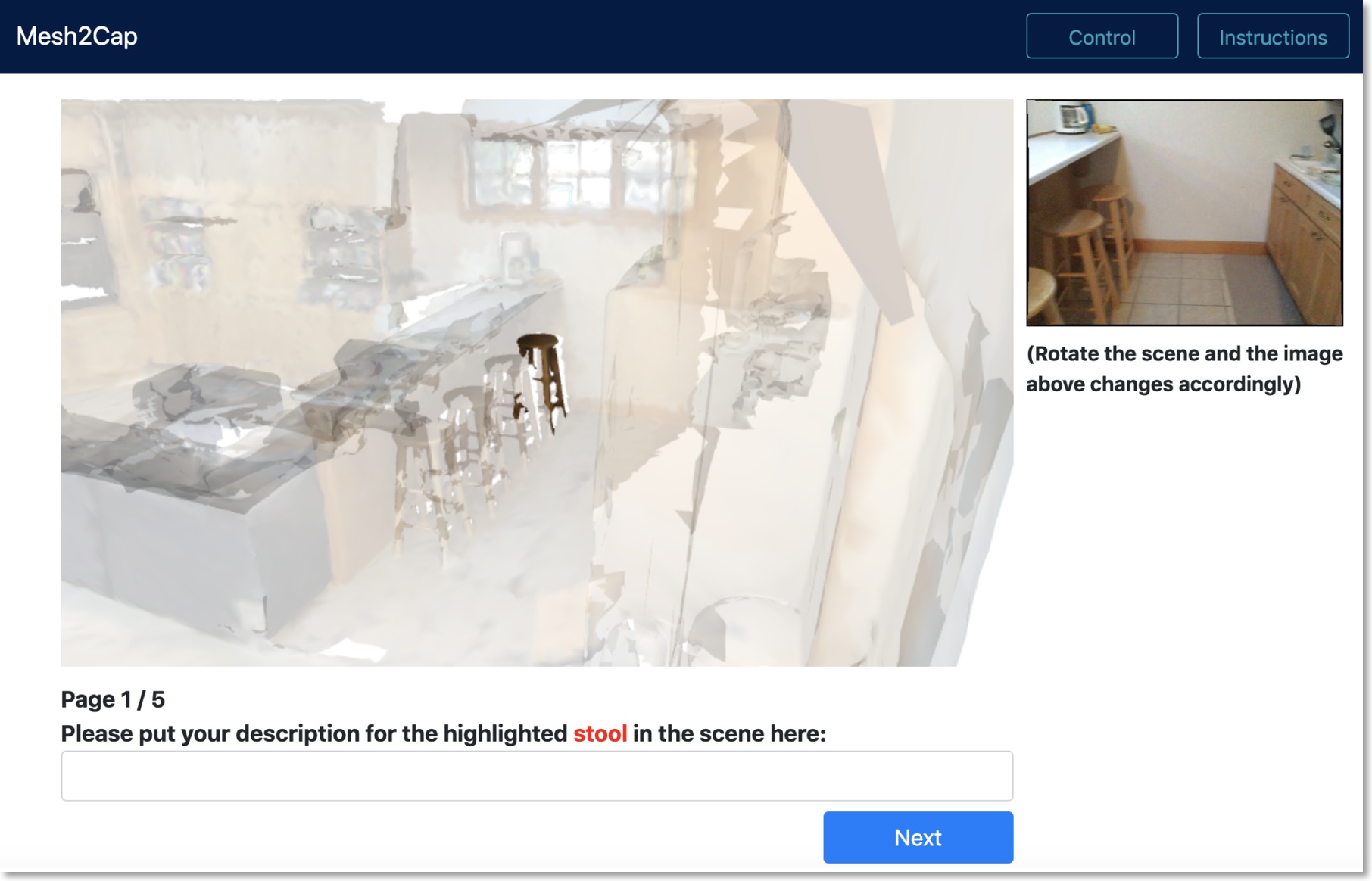}
        \caption{Annotation interface for Amazon Mechanical Turk workers used to create the ScanRefer dataset.}
        \label{fig:annotation}
    \end{subfigure}
    
    \begin{subfigure}{0.9\textwidth}
        \centering
        \includegraphics[width=\textwidth]{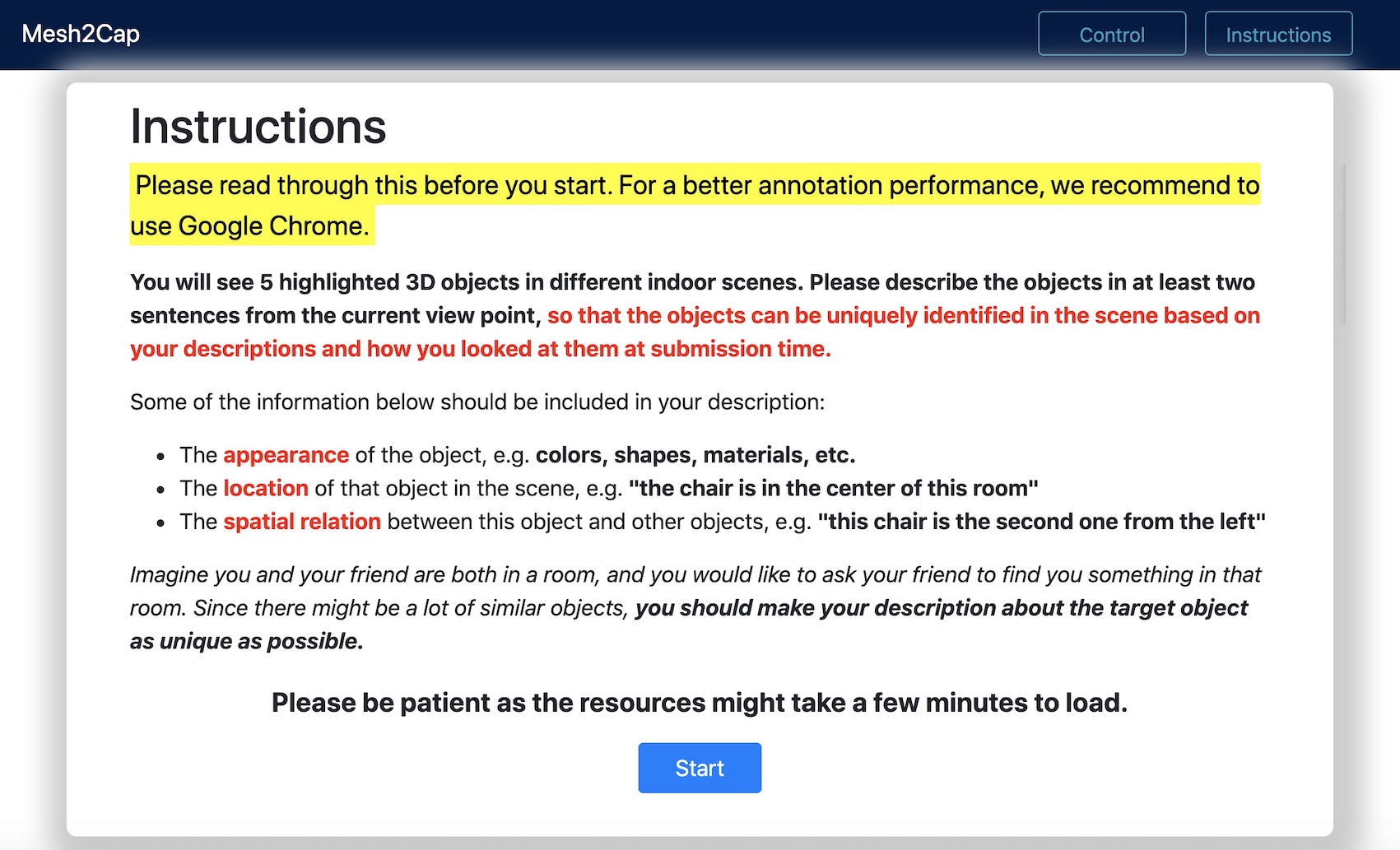}
        \caption{Annotation instructions shown to the Amazon Mechanical Turk workers.}
        \label{fig:annotation_instructions}
    \end{subfigure}
    \caption{(a) Our web-based annotation interface: annotators are requested to describe a batch of 5 target objects. The viewpoint can be adjusted by the user while the image on the right is chosen based on the camera view. (b) Screenshot of the instructions for the Amazon Mechanical Turk workers before providing descriptions for objects.}
\end{figure*}

\begin{figure*}[t!]
    \centering
    \begin{subfigure}{0.9\textwidth}
        \centering
        \includegraphics[width=\textwidth]{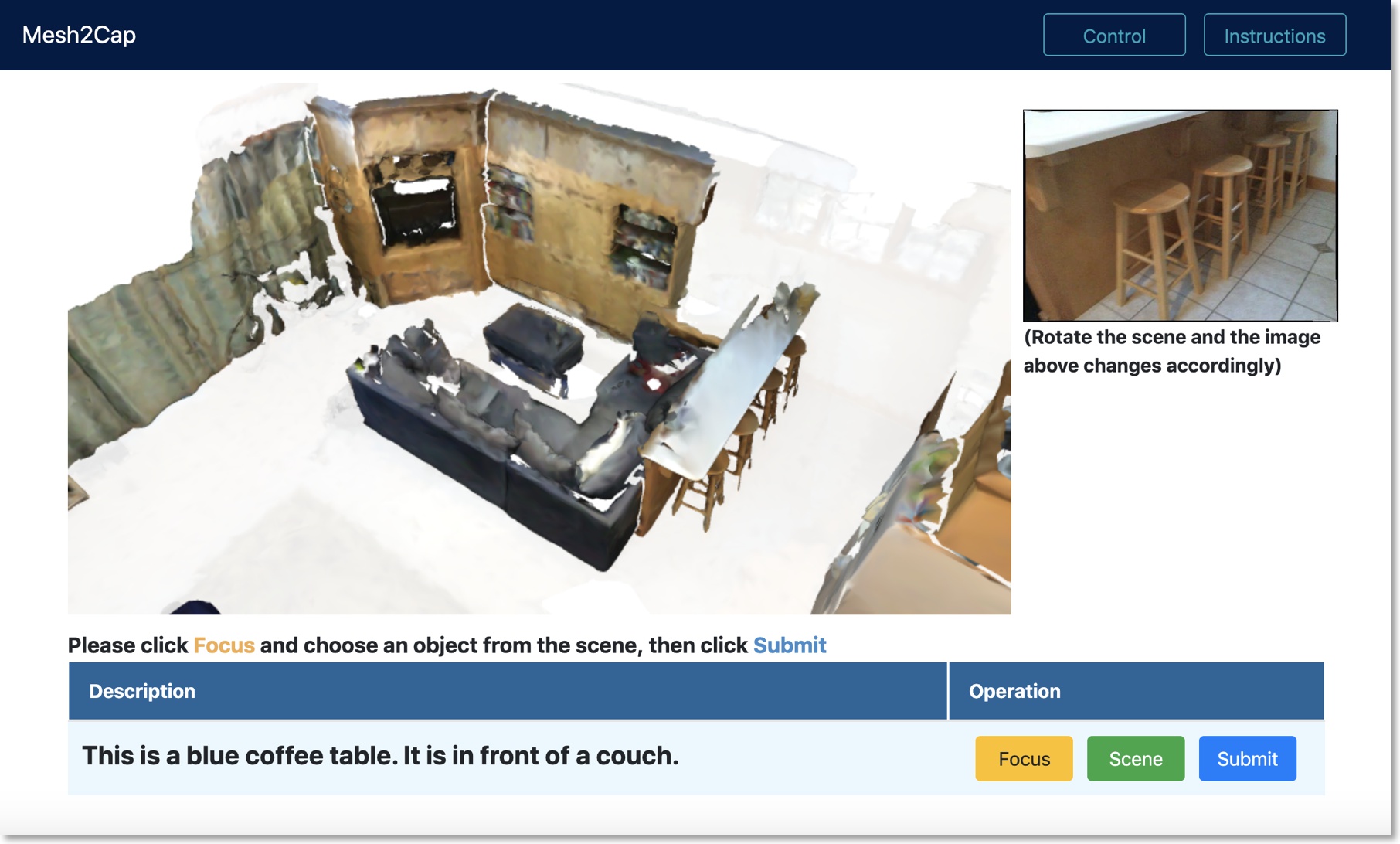}
        \caption{Verification interface used by trained student verifiers in order to verify each annotation done earlier by the annotation Amazon Mechanical Turk workers.}
        \label{fig:verification}
    \end{subfigure}
    \begin{subfigure}{0.9\textwidth}
        \centering
        \includegraphics[width=\textwidth]{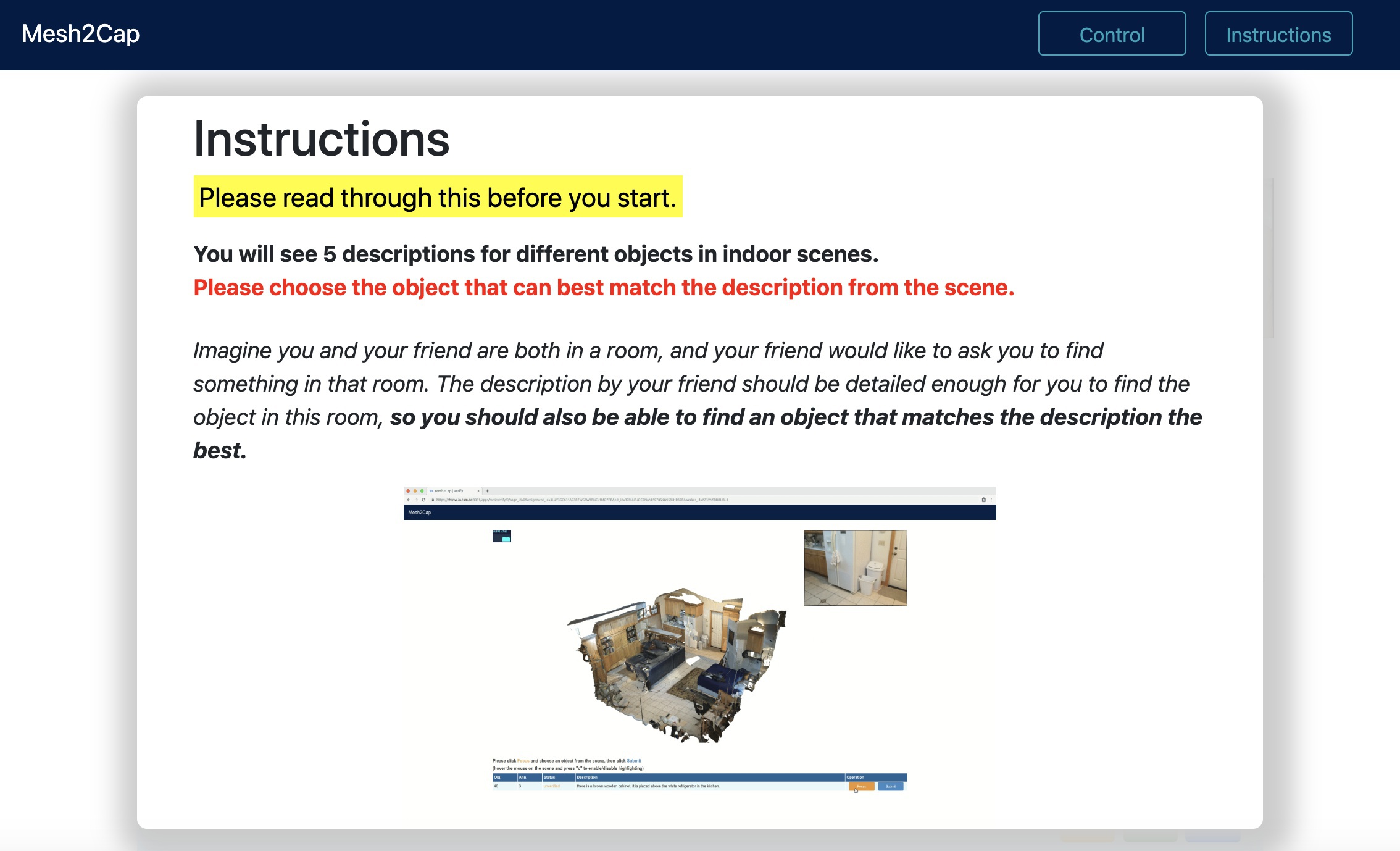}
        \caption{Verification instructions shown to the trained student verifiers.}
        \label{fig:verification_instructions}
    \end{subfigure}
    \caption{(a) Our web-based verification interface: verifiers are asked to select objects that match the provided descriptions from the collection step. The ambiguous descriptions, which can be used to match multiple objects in the scene, are excluded from the final dataset. (b) Screenshot of the instructions that the trained verifiers have to go through before starting the verification.}
\end{figure*}

\subsubsection{Annotation}
\label{sec:annotation}

We deploy our web-based annotation application on Amazon Mechanical Turk (AMT) to collect object descriptions in the reconstructed RGB-D scans, as shown in Fig.~\ref{fig:annotation}. To ensure that the initial descriptions are written in proper English, we restrict the workers to be from the United States, the United Kingdom, Canada, and Australia. The workers are asked to finish a batch of 5 description tasks within a time limit of 2 hours once the assignment is accepted on AMT. To ensure the descriptions are diverse and linguistically rich, we require that each description consists of at least two sentences.
Before the annotation task begins, the AMT workers are also presented with the instructions shown in Fig.~\ref{fig:annotation_instructions}. We request that the workers provide the following information in the descriptions:

\begin{itemize}
\item The appearance of the object such as shape, color, material and so on.
\item The location of that object in the scene, e.g., ``the chair is in the center of this room".
\item The relative position to other objects in the scene, for instance, ``this chair is the second one from the left".
\end{itemize}

\subsubsection{Verification}
\label{sec:verification}

After collecting the descriptions from AMT, we do a quick inspection of the descriptions and manually filter and reject obvious bad descriptions before we start the verification process. We then verify the collected object descriptions by recruiting trained students to perform the verification task on our WebGL-based application, as shown in Fig.~\ref{fig:verification}. To ensure that the descriptions provided are discriminative (e.g., can pick out which one of the chairs is being described), the verifiers are asked to select the objects in the scene that match the descriptions the best. The verifiers are also asked to fix any spelling and wording issues, e.g., ``hair'' instead of ``chair'', and submit the corrected descriptions to our database. To guide the trained verifiers, we provide the verification instructions as shown in Fig.~\ref{fig:verification_instructions}.

\section{Additional Implementation Details}
\label{sec:implemention}

\begin{figure*}[!t]
    \centering
    \includegraphics[width=0.6\textwidth]{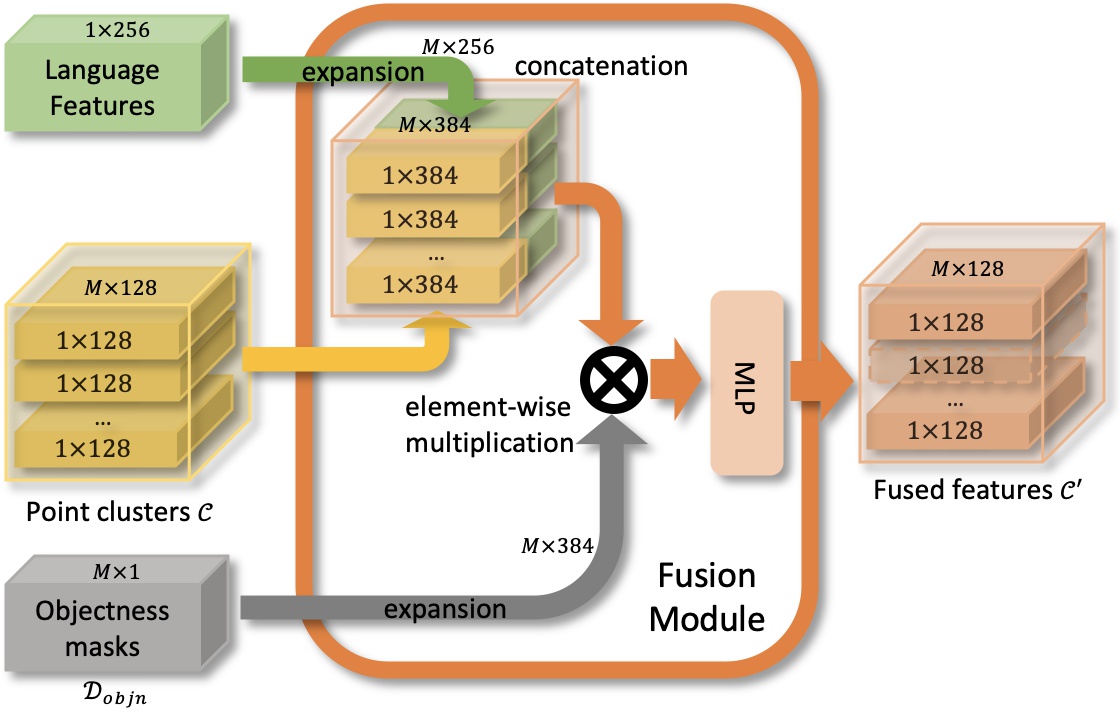}
    \caption{The fusion module takes as input the aggregated point clusters, the language embeddings, and the predicted objectness masks. It first concatenates the point clusters with the expanded language features as the raw fused features, of which the invalid ones will be masked out by the predicted objectness masks. Finally, a multi-layer perceptron takes in the raw fused features and outputs the final fused multimodal point features.}
    \label{fig:fusion module}
\end{figure*}

\subsection{Fusion Module}

Fig.~\ref{fig:fusion module} shows the feature fusion process in our localization pipeline. Concretely, the fusion module first concatenates the point clusters $C={c_i} \in \mathcal{R}^{M \times 128}$ and expanded language embedding $E={e'} \in \mathcal{R}^{M \times 256}$, then multiply the expanded objectness mask $D_{objn}' \in \mathcal{R}^{M \times 384}$ to filter out invalid object proposals. A multi-layer perceptron maps the filtered feature maps into the final fused features $\mathcal{C}' \in \mathcal{R}^{M \times 128}$ as the output of the fusion module.

\begin{table*}[!t]
    \centering
    \resizebox{\textwidth}{!}{%
        \begin{tabular}[width=\textwidth]{l*{18}{c}|c}
            \toprule
            & cab. & bed & chair & sofa & tabl. & door & wind. & bkshf. & pic. & cntr. & desk & curt. & fridg. & showr. & toil. & sink & bath. & others & mAP \\
            \midrule
            {[a]} & 4.77 & 85.51 & 64.42 & 72.74 & 30.39 & 11.17 & 6.62 & 17.32 & 0.35 & 2.16 & 35.79 & 7.80 & 16.69 & 16.96 & 76.74 & 16.77 & 69.57 & 5.68 & 30.08 \\
            \midrule
            {[b]} & 9.93 & \textbf{88.43} & 67.12 & 69.44 & \textbf{39.76} & 12.20 & 5.11 & 20.27 & 0.02 & 9.27 & 41.52 & 16.10 & \textbf{30.79} & 5.77 & 77.32 & 14.93 & 61.02 & 7.82 & 32.05 \\
            {[c]} & 7.01 & 88.01 & 67.13 & 73.69 & 32.87 & 12.36 & 9.01 & 17.61 & 0.31 & 9.27 & 44.78 & 16.25 & 20.29 & 3.55 & 76.50 & 12.33 & 72.24 & 8.08 & 31.74 \\
            {[d]} & 11.16 & 87.20 & \textbf{70.58} & 75.17 & 36.76 & 11.47 & 6.72 & 13.40 & 1.09 & 7.08 & 48.38 & 11.64 & 19.96 & 4.29 & 85.29 & \textbf{18.20} & 72.83 & \textbf{10.74} & 32.89 \\
            {[e]} & 7.22 & 87.72 & 67.24 & 72.42 & 33.66 & 11.55 & 8.80 & 20.16 & 0.14 & \textbf{9.82} & 46.07 & 15.91 & 22.48 & 2.67 & 77.82 & 13.17 & 68.14 & 8.01 & 31.83 \\
            {[f]} & \textbf{12.74} & 83.91 & 69.94 & 72.17 & 36.11 & \textbf{13.38} & 8.42 & 17.52 & \textbf{1.99} & 6.58 & \textbf{46.65} & \textbf{17.65} & 24.04 & \textbf{31.30} & 75.99 & 10.31 & 61.92 & 9.78 & \textbf{33.36} \\
            \midrule
            {[g]} & 10.53 & 84.00 & 63.48 & \textbf{75.27} & 30.62 & 7.78 & 8.45 & 18.08 & 1.18 & 5.47 & 39.27 & 10.14 & 18.83 & 8.93 & 69.99 & 9.36 & \textbf{75.59} & 7.97 & 30.27 \\
            {[h]} & 11.11 & 85.63 & 67.81 & 71.04 & 34.96 & 9.54 & 6.22 & 16.37 & 1.67 & 6.28 & 36.07 & 12.93 & 17.40 & 7.46 & 68.74 & 11.77 & 65.69 & 7.71 & 29.91 \\
            {[i]} & 10.72 & 86.71 & 69.86 & 72.77 & 32.60 & 16.33 & 8.16 & 19.64 & 1.14 & 7.08 & 42.21 & 14.31 & 22.99 & 6.92 & \textbf{86.09} & 8.06 & 65.51 & 8.79 & 32.22 \\
            {[j]} & 9.76 & 87.93 & 65.93 & 72.59 & 31.60 & 9.48 & 9.05 & \textbf{23.86} & 0.37 & 6.69 & 42.22 & 13.86 & 21.42 & 16.35 & 80.41 & 12.30 & 57.80 & 7.40 & 31.61 \\
            {[k]} & 8.92 & 88.20 & 70.37 & 73.93 & 32.89 & 10.54 & \textbf{9.21} & 14.05 & 0.48 & 6.91 & 44.74 & 6.54 & 17.76 & 27.64 & 81.18 & 12.86 & 62.40 & 9.06 & 32.09 \\
            \bottomrule
        \end{tabular}
    }
    \caption{Object detection results measured using mean average precision (mAP) at IOU of 0.5 for the 18 difference classes for [a] VoteNet~\citep{qi2019deep}, [b] Ours (xyz), [c] Ours (xyz+rgb), [d] Ours (xyz+rgb+normals), [e] Ours (xyz+multiview), [f] Ours (xyz+multiview+normals), [g] Ours (xyz+lobjcls), [h] Ours (xyz+rgb+lobjcls), [i] Ours (xyz+rgb+normals+lobjcls), [j] Ours (xyz+multiview+lobjcls), [k] Ours (xyz+multiview+normals+lobjcls). Training with  point normals (compare rows [d,f] to rows [c,e]) and multiview features (compare rows [e,f] to rows [c,d]) clearly leads to better performance. As expected, models with the language-based object classifier (rows [g-k]) does not results in better object detection compared to  models without such a module (rows [b-f]).}
    \label{tab:detection}
\end{table*}

\section{Additional quantitative analysis}
\label{sec:supquant}

\subsection{Object Detection Results}

In order to evaluate the 3D object detection, we conduct ablations of our architecture with different point cloud features as well as ablating the inclusion of the language-based object classifier (see Tab.~\ref{tab:detection}).
We also compare against the the object detection results of VoteNet~\citep{qi2019deep}.  
We use the mean average precision (mAP) thresholded by IoU value $0.5$ as our evaluation metric and examine the object detection results for different object categories. 
We exclude structural objects such as ``Floor'' and ``Wall''. We group all categories which are not in the ScanNet benchmark categories \citep{dai2017scannet} including ``Otherfurnitures'', ``Otherstructure'', and ``Otherprop'' into the ``Others'' category  in our evaluation.  
Note that the ``Others'' category in our evaluation includes additional types of objects, such as ``Pillow'' and ``Keyboard'', with respect to those in the ``Otherfurniture'' category of the ScanNet benchmark.

While our 3D object detector is robust in identifying and separating out instances of large objects that are typically placed away from walls (e.g., bed, chair, sofa, toilet, bathtub), it is not as reliable at identifying instances of flat objects (e.g., picture, window, door) and objects with unclear instance boundaries (e.g., cabinet, shelving) and smaller objects (e.g., sink, others). 
Overall, our best 3D object detector only achieves a mAP of $33\%$, suggesting that improving 3D object detection, especially better instance detection for the ``other'' category, is a key challenge in our task of localizing objects in 3D using natural language.

As shown in Tab.~\ref{tab:detection}, including point normals as extra point features (rows [d,f]) in training increases the detection results when compared to the models trained without the normals (rows [c,e]). 
Also, training with extracted high-level color features from the multi-view images (rows [e,f]) also produces better detection results compared with the results from models trained with just the raw RGB values (rows [c,d]). 
Note that networks equipped with the language-based object classifier (rows [g-k]) fail to produce better detection results compared to the ones without the extra language classifier module (rows [b-f]).  
This behavior is expected as the description provides additional information which helps to differentiate between objects of the same category; but it has no information for helping with object detection.

\subsection{Training and Evaluation Variance}

\begin{table*}[!t]
    \centering

    \begin{tabular}{l*{3}{r}}
        \toprule
        & unique & multiple & overall \\
        random seed & Acc@0.5 & Acc@0.5 & Acc@0.5 \\
        \midrule
        \updated{2} & \updated{46.83} & \updated{20.57} & \updated{25.66} \\
        \updated{4} & \updated{47.96} & \updated{19.45} & \updated{24.98} \\
        \updated{8} & \updated{45.96} & \updated{20.05} & \updated{25.07} \\
        \midrule
        \updated{standard deviation} & \updated{0.82} & \updated{0.46} & \updated{0.30} \\
        \updated{mean} & \updated{46.92} & \updated{20.02} & \updated{25.23} \\
        \bottomrule
    \end{tabular}

    \caption{Variance between training runs. We train our model (xyz+rgb+lobjcls) with three different random seeds ($2$, $4$, $8$) and evaluate the trained model using a fixed random seed $42$.  We have a training stddev of $0.30$. 
    }
    \label{tab:train_variance}
\end{table*}

\begin{table*}[!t]
    \centering
    \begin{tabular}{l*{3}{r}}
        \toprule
        & unique & multiple & overall \\
        random seed & Acc@0.5 & Acc@0.5 & Acc@0.5 \\
        \midrule
        \updated{42} & \updated{48.89} & \updated{22.24} & \updated{27.40} \\
        \updated{2} & \updated{49.28} & \updated{22.05} & \updated{27.34} \\
        \updated{4} & \updated{48.68} & \updated{21.56} & \updated{26.82} \\
        \updated{8} & \updated{48.29} & \updated{21.99} & \updated{27.09} \\
        \updated{16} & \updated{50.35} & \updated{21.42} & \updated{27.03} \\
        \updated{32} & \updated{49.55} & \updated{21.75} & \updated{27.14} \\
        \updated{64} & \updated{49.61} & \updated{22.25} & \updated{27.56} \\
        \updated{128} & \updated{49.28} & \updated{21.57} & \updated{26.95} \\
        \updated{256} & \updated{49.88} & \updated{21.98} & \updated{27.39} \\
        \updated{512} & \updated{47.29} & \updated{21.99} & \updated{28.12} \\
        \midrule
        \updated{standard deviation} & \updated{0.87} & \updated{0.29} & \updated{0.37} \\
        \updated{mean} & \updated{49.11} & \updated{21.88} & \updated{27.28} \\
        \bottomrule
    \end{tabular}
    \caption{Variance between evaluation runs due to the random sampling of points in the VoteNet~\citep{qi2019deep}. We train our model (xyz+multiview+normal+lobjcls) with the a fixed random seed of $42$ and evaluate the trained model using $10$ different random seeds as shown in the first column. We have a evaluation stddev of $0.37$.
    }
    \label{tab:eval_variance}
\end{table*}

\updated{Since there is a random sampling of 40,000 points from the original point cloud in the VoteNet~\citep{qi2019deep} detection backbone, we conduct experiments to measure the training and evaluation variance across multiple runs. 
As shown in Tab.~\ref{tab:train_variance} and Tab.~\ref{tab:eval_variance}, due to random sampling, there is a stddev of $0.30$ across training runs and a stddev of $0.37$ across evaluation runs. For more reliable results, we average the results of 5 evaluation runs with different random seeds when using VoteNet.}

\subsection{Additional Ablation Study}

\begin{table*}[!t]
    \centering
    \resizebox{\columnwidth}{!}{
        \begin{tabular}{l*{6}{c}}
            \toprule
            & \multicolumn{2}{c}{unique} & \multicolumn{2}{c}{multiple} & \multicolumn{2}{c}{overall} \\
            & Acc@0.25 & Acc@0.5 & Acc@0.25 & Acc@0.5 & Acc@0.25 & Acc@0.5 \\
            \midrule
            \updated{Ours (semantic labels)} & \updated{61.60} & \updated{39.04} & \updated{28.26} & \updated{18.98} & \updated{34.72} & \updated{21.88} \\
            \updated{Ours (object names)} & \updated{70.53} & \updated{44.69} & \updated{32.34} & \updated{20.33} & \updated{39.75} & \updated{25.05} \\
            \updated{Ours (first sentences)} & \updated{73.52} & \updated{46.60} & \updated{33.71} & \updated{21.20} & \updated{41.44} & \updated{26.12} \\
            \updated{Ours (whole descriptions)} & \textbf{76.33} & \textbf{53.51} & \textbf{32.73} & \textbf{21.11} & \textbf{41.19} & \textbf{27.40} \\
            \bottomrule
        \end{tabular}
    }
    \caption{Ablation study with different input lengths. We measure the percentages of predictions whose IoU with the ground truth boxes are greater than 0.25 and 0.5. Unique means that there is only a single object of its class in the scene. Obviously, the richer information the descriptions contain, the better our localization pipeline performs.}
    \label{tab:length}
\end{table*}

In Tab.~\ref{tab:length}, we examine what happens when we feed different language inputs into our pipeline.

\mypara{Does our method really learn from the full descriptions?}
To evaluate the impact of information from the full descriptions versus just the identification of the type of object to locate, we compare using the full description as input versus using the semantic label or the object name as the input.  
For example, for a target object ``trash can'' with the description \textit{This is a short trash can. 
It is in front of a taller trash can.}, we input ``trash can'' as the object name and ``others'' as the semantic label (see Sec.~\ref{sec:statistics} for list of semantic classes). 
The results in Tab.~\ref{tab:length} show that using the full descriptions improves the localization performance compared to using just the semantic labels as input. 
Comparing the performance of using semantic labels and object names, we see that inputting the semantic labels helps with the performance in the ``unique'' scenarios where there is only one object from a certain category, but suffers in the ``multiple'' scenarios where more information is needed to distinguish between objects that are grouped into the same broad category (e.g., ``trash can'' and ``laptop'' would both be categorized as ``other'', and ``armchair'' would provide more information than just the coarse semantic label ``chair'').

\mypara{Are the first sentences enough for the task?}
Since we deliberately collect at least two sentences as descriptions for the objects to ensure the richness of information, we also conduct experiments to show that the full description (with potentially multiple sentences) result in better performance than using only the first sentences. As Tab.~\ref{tab:length} shows, the model trained on longer descriptions performs better than the one trained just on the first sentences.

\begin{figure*}[!t]
    \centering
    \includegraphics[width=0.99\textwidth]{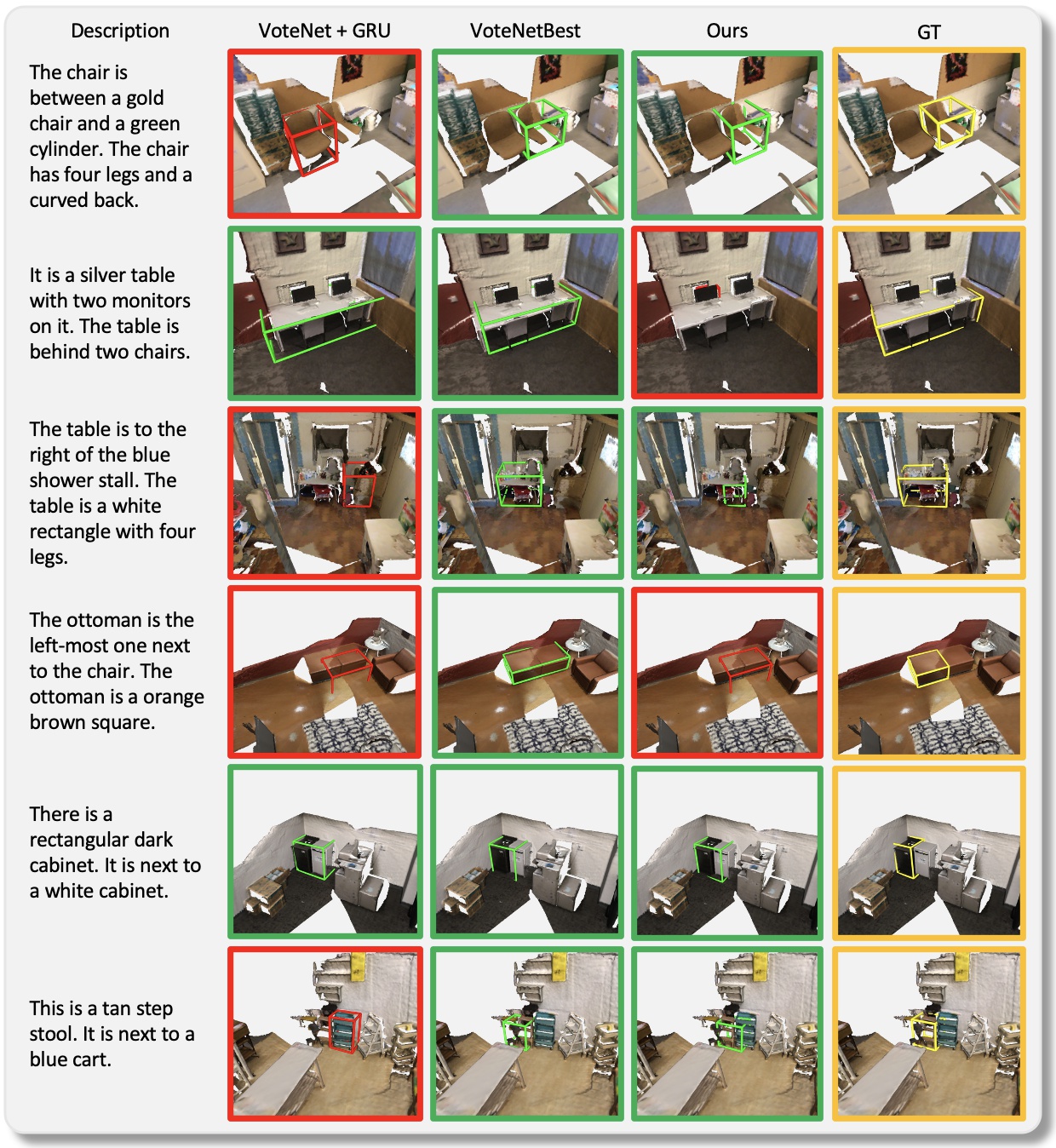}
    \caption{Additional qualitative analysis comparing our method with VoteNet~\citep{qi2019deep}+GRU and VoteNetBest.}
    \label{fig:extra_votenet}
\end{figure*}

\begin{figure*}[!t]
    \centering
    \includegraphics[width=0.99\textwidth]{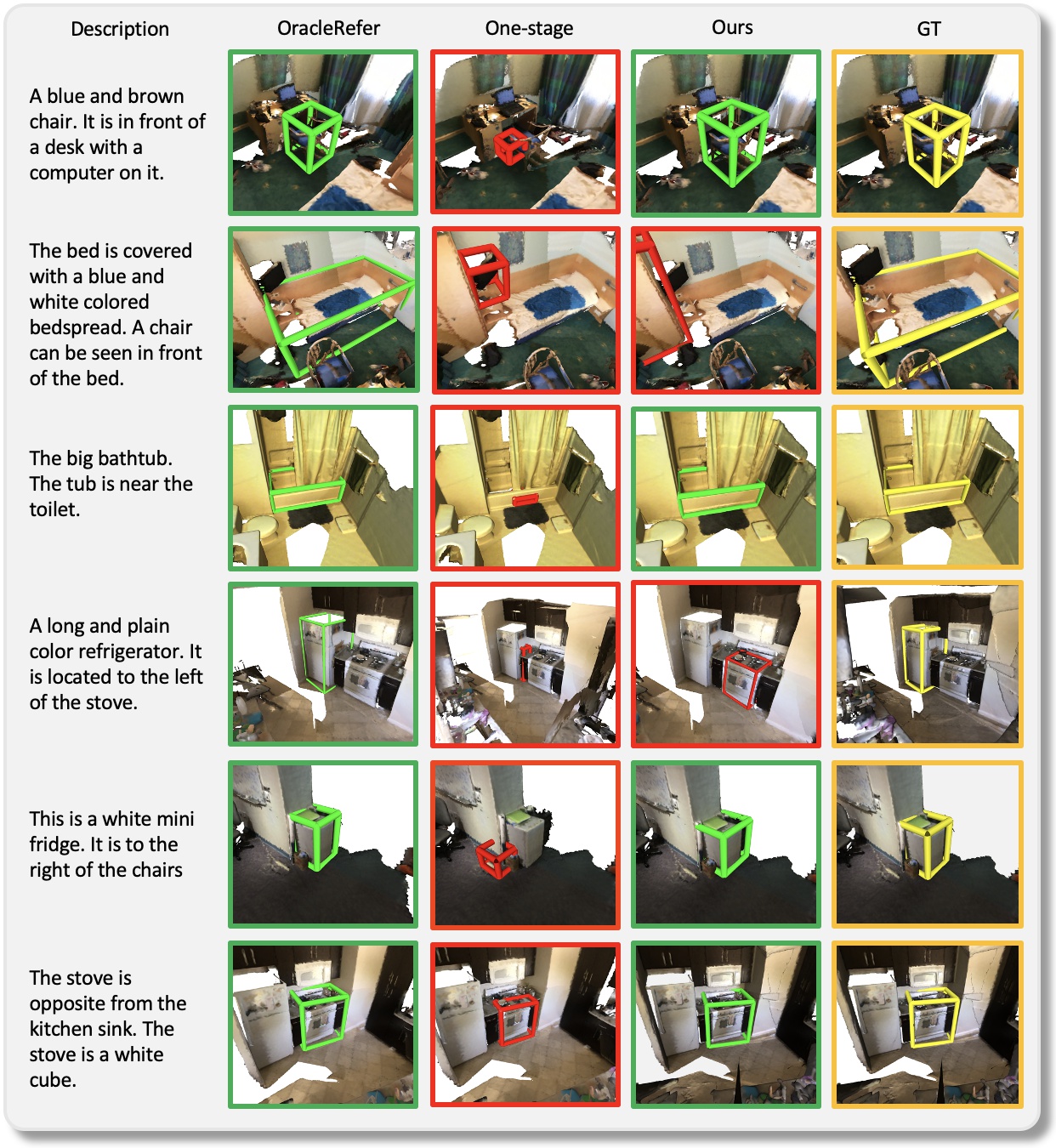}
    \caption{Additional qualitative analysis in the ``unique'' scenarios where there is only one object from a certain category. Our method is capable of localizing the target object in a 3D indoor scene with the help of the free-form description.}
    \label{fig:extra_unique}
\end{figure*}

\begin{figure*}[!t]
    \centering
    \includegraphics[width=0.99\textwidth]{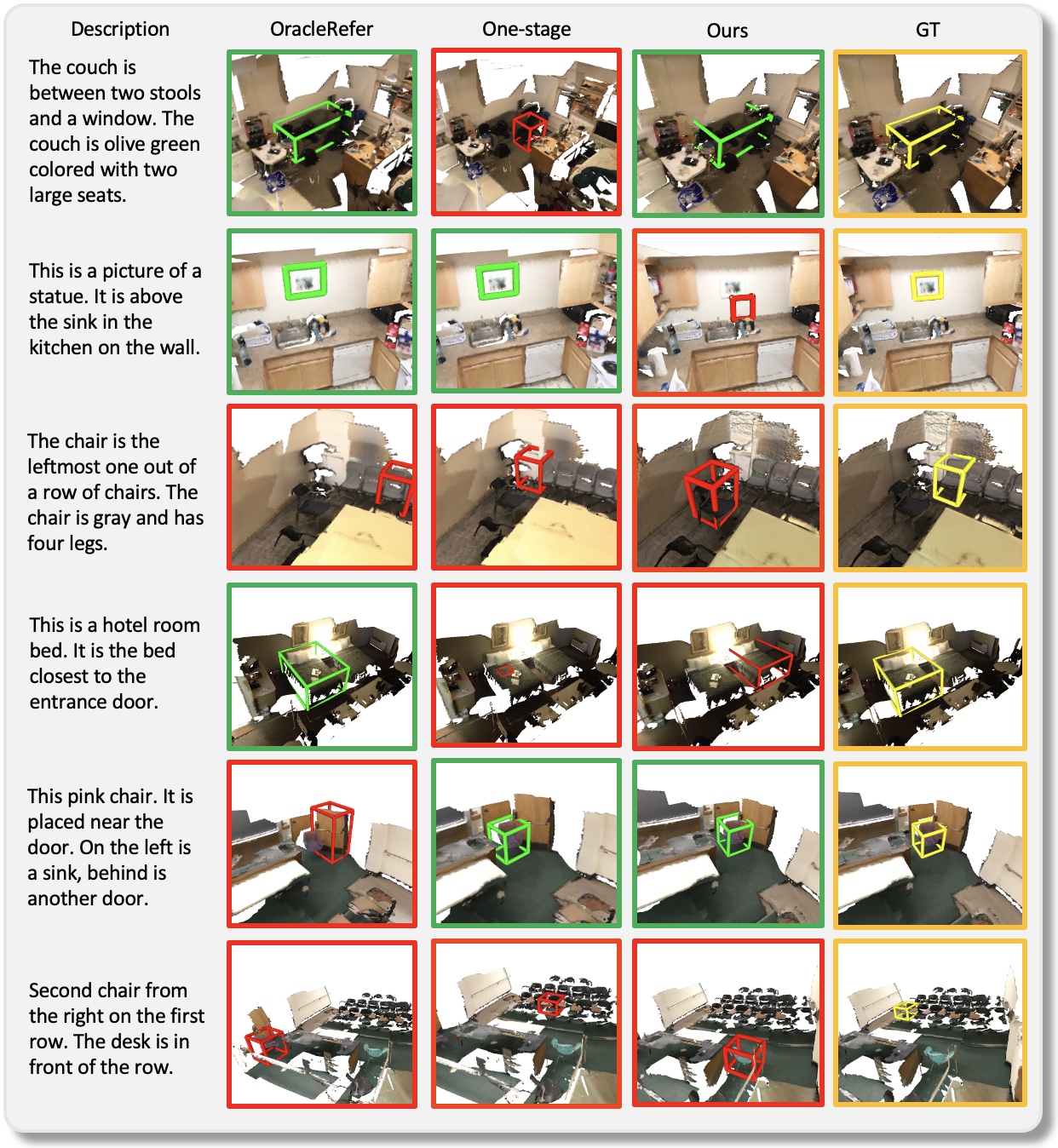}
    \caption{Additional qualitative analysis for the ``multiple'' subset where there are multiple objects with the same category as the target objects. While our methods can correctly localize the target object in some cases (rows 1,5), it often fails due to the limited accuracy of the object detector (row 2) or difficulty disambiguating between multiple instances (rows 3,4,6).}
    \label{fig:extra_multiple_1}
\end{figure*}

\begin{figure*}[!t]
    \centering
    \includegraphics[width=0.99\textwidth]{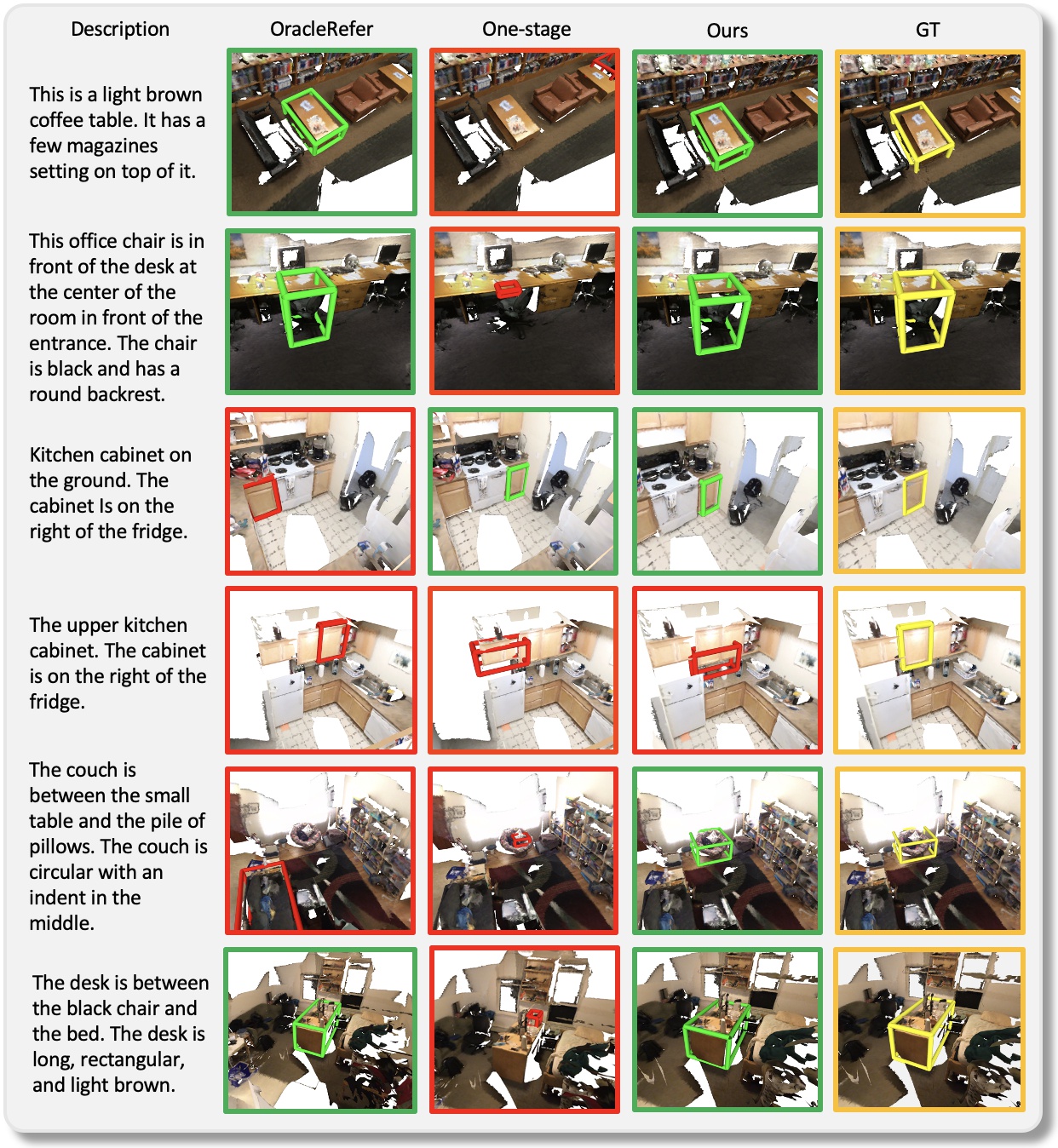}
    \caption{Additional qualitative analysis for the ``multiple'' subset where there are multiple objects with the same category as the target objects. While our methods can correctly localize the target object in some cases (rows 1-3,5-6), it can fail due to the limited accuracy of the object detector and difficulty handling spatial relations (rows 4).}
    \label{fig:extra_multiple_2}
\end{figure*}

\section{Additional Qualitative Analysis}
\label{sec:supqual}

We present additional examples of localization results by our method and the baselines for further qualitative analysis. 

\updated{\mypara{Qualitative results comparing VoteNet~\citep{qi2019deep}+GRU and VoteNetBest with out method}
We show more qualitative results in Fig.~\ref{fig:extra_votenet} to display the difference in performance between these three methods. As shown in the first column in Fig.~\ref{fig:extra_votenet}, using a pretrained VoteNet~\citep{qi2019deep} detection backbone provides reasonable bounding box around objects, but still performs slightly worse than our method where we train the detection backbone and localization module in an end-to-end fashion (see the third column "ours").
}

\mypara{More qualitative examples comparing OracleRefer and One-stage (with 2D to 3D backprojection) with our method}
To illustrate the difference in performance between the methods, we provide more qualitative results.
We split the localization results into ``unique'' (Fig.~\ref{fig:extra_unique}) and ``multiple'' (Fig.~\ref{fig:extra_multiple_1} \& Fig.~\ref{fig:extra_multiple_2}) subsets.  
As shown in Fig.~\ref{fig:extra_unique}, for the ``unique'' subset, our method is able to identify and localize the object. 
In contrast, the 2D method (One-Stage), is able to identify the rough location of the object, but the backprojected 3D bounding box does not match the ground truth very well.
For the ``multiple'' subset, there are challenging cases where our method fails to localize the target object.
Fig.~\ref{fig:extra_multiple_1} and~\ref{fig:extra_multiple_2} show that our method is able to localize objects correctly (Fig.~\ref{fig:extra_multiple_1} rows 1,5, Fig.~\ref{fig:extra_multiple_2} rows 1-3,5-6) even when there are other objects of the same category in the scene.  Our method is sometimes limited by the accuracy of the object detector, which tends to produce inaccurate bounding boxes for small objects such as pictures (Fig.~\ref{fig:extra_multiple_1} row 2).
This indicates that the object detection  can still be improved.  Our method also has trouble disambiguating between objects based on spatial relations (Fig.~\ref{fig:extra_multiple_1} rows 3-4,6).
For instance, for comparative phrases (e.g., ``leftmost'' or ``rightmost'') or counting (e.g., ``the second one from the left''), the model fails to pick out the correct object (Fig.~\ref{fig:extra_multiple_1} rows 4).

\end{document}